%% file: main.tex
\PassOptionsToPackage{table}{xcolor}
\documentclass[sigconf]{acmart}
\AtBeginDocument{%
  }

\copyrightyear{2026}
\acmYear{2026}
\setcopyright{cc}
\setcctype{by}
\acmConference[MM '26]{Proceedings of the 34th ACM International Conference on Multimedia}{November 10--14, 2026}{Rio de Janeiro, Brazil}
\acmBooktitle{Proceedings of the 34th ACM International Conference on Multimedia (MM '26), November 10--14, 2026, Rio de Janeiro, Brazil}
\acmDOI{10.1145/3767308.3835471}
\acmISBN{979-8-4007-2213-4/2026/11}

\input{preamble}

\title{DEFUSE: Generalizable Backdoor Defense for Self-Supervised Encoders with Generative Priors}

\author{Tuo Chen}
\affiliation{%
  \department{School of Cyber Science and Engineering}
  \institution{Southeast University}
  \city{Nanjing}
  \country{China}}
\affiliation{%
  \institution{Ant Group}
  \city{Hangzhou}
  \country{China}}
\email{tchen@seu.edu.cn}

\author{Jie Gui}
\authornote{Corresponding authors.}
\affiliation{%
  \department{School of Cyber Science and Engineering}
  \institution{Southeast University}
  \city{Nanjing}
  \country{China}}
\affiliation{%
  \institution{Purple Mountain Laboratories}
  \city{Nanjing}
  \country{China}}      
\email{guijie@seu.edu.cn}

\author{Minjing Dong}
\affiliation{%
  \department{Department of Computer Science}
  \institution{City University of Hong Kong}
  \city{Hong Kong}
  \country{China}}
\email{minjdong@cityu.edu.hk}

\author{Lanting Fang}
\affiliation{%
  \institution{Beijing Institute of Technology}
  \city{Beijing}
  \country{China}}
\email{lantingf@outlook.com}

\author{Ju Jia}
\affiliation{%
  \department{School of Cyber Science and Engineering}
  \institution{Southeast University}
  \city{Nanjing}
  \country{China}}
\email{jiaju@seu.edu.cn}

\author{Benlei Cui}
\affiliation{%
  \institution{Alibaba Group}
  \city{Hangzhou}
  \country{China}}
\email{cubenlei.cbl@alibaba-inc.com}

\author{Jian Liu}
\authornotemark[1]
\affiliation{%
  \institution{Ant Group}
  \city{Hangzhou}
  \country{China}}
\email{rex.lj@antgroup.com}

\renewcommand{\shortauthors}{Tuo Chen et al.}

\begin{document}

\begin{abstract}
\input{sec/0_abstract}
\end{abstract}



\maketitle

\input{sec/intro}
\input{sec/related_work}
\input{sec/method}

\input{sec/experiment}

\section{Conclusion}
We present DEFUSE, a generalizable backdoor detection framework for self-supervised encoders. By projecting suspicious representations back to the image space and evaluating semantic consistency, DEFUSE identifies anomalous representations without requiring prior knowledge about the victim encoder or the attack strategy. Extensive experiments across visual SSL and vision-language encoders demonstrate that DEFUSE delivers strong and consistent detection performance under diverse evaluation settings. Overall, these results validate the effectiveness and generalizability of our framework, highlighting the potential of generative priors for securing self-supervised encoders in practice.

\clearpage
\bibliographystyle{ACM-Reference-Format}
\bibliography{zotero}

\clearpage
\appendix
\setcounter{section}{0}
\setcounter{figure}{0}
\setcounter{table}{0}
\setcounter{equation}{0}
\renewcommand{\thesection}{S\arabic{section}}
\renewcommand{\thesubsection}{\thesection.\arabic{subsection}}
\renewcommand{\thefigure}{S\arabic{figure}}
\renewcommand{\thetable}{S\arabic{table}}
\renewcommand{\theequation}{S\arabic{equation}}
\input{sec/appe}

\end{document}

%% file: preamble.tex
\usepackage{xcolor}
\usepackage{colortbl}
\usepackage{graphicx}
\usepackage{booktabs}
\usepackage{multirow}
\usepackage{nicefrac}
\usepackage{microtype}
\usepackage{amsmath,amsfonts,amsthm}
\usepackage{bm}
\usepackage{dsfont}

\usepackage{subcaption}
\usepackage{placeins}
\usepackage{cuted}
\usepackage{algorithm}
\usepackage{algorithmic}
\usepackage{newfloat}
\usepackage{listings}
\usepackage{pifont}

\definecolor{lightgray}{gray}{0.9}

\DeclareCaptionStyle{ruled}{labelfont=normalfont,labelsep=colon,strut=off}
\floatstyle{ruled}
\newfloat{listing}{tb}{lst}
\floatname{listing}{Listing}

%% file: sec/0_abstract.tex
Self-supervised learning (SSL) encoders are vulnerable to backdoor attacks, posing threats to both visual SSL encoders and vision-language encoders. Existing defenses are typically designed for only one of these paradigms and rely on restrictive assumptions such as access to uninfected in-distribution data or precomputed pseudo-labels, which are difficult to satisfy in practice.
To address these limitations, we propose DEFUSE, a generalizable backdoor detection framework for SSL encoders. Inspired by Bayesian posterior inference, we reformulate backdoor detection as a representation-conditioned image likelihood estimation problem parameterized by a conditional diffusion generative model. Uninfected representations tend to yield semantically consistent reconstructions, whereas backdoored ones are more likely to be mapped to the attacker's target class or semantically meaningless images, deviating from the original semantics and thereby exposing the backdoor.
However, we find that the exact likelihood is intractable, because highly abstracted representations discard the low-level information necessary for pixel-faithful reconstruction. We therefore relax the objective to semantic reconstruction and evaluate it in a well-separated representation space provided by a reference encoder. Rather than training from scratch, we fine-tune a pretrained diffusion model, leveraging its generative prior to map data onto the natural image manifold while preserving semantic content.
Extensive experiments demonstrate that DEFUSE substantially outperforms existing detectors across diverse attack settings, generalizing to both visual SSL and vision-language encoders. Notably, our method greatly reduces the reliance on prior knowledge about the victim encoder or the attack strategy.
The source code is available at \url{https://github.com/jsrdcht/DEFUSE}.

%% file: sec/intro.tex
\section{Introduction}

Self-supervised learning (SSL) has achieved significant progress in recent years. Trained on internet-scale corpora spanning images, videos, and image-text pairs, SSL has become a standard paradigm for pretraining visual encoders. This paradigm extracts richer information from data than supervised learning, allowing model providers to scale up model capacity and train large foundation models~\cite{he2022Masked,radford2021Learning,bai2023qwen}.
For instance, CLIP~\cite{radford2021Learning} is pretrained by OpenAI on 400M image-text pairs collected from the Internet, which endows it with remarkable zero-shot transfer ability across a wide range of downstream tasks. Model providers can monetize their encoders by deploying them as cloud services and offering APIs to users~\cite{liu2022PoisonedEncoder}. Users query these APIs to obtain feature vectors, which then serve as the backbone for various downstream tasks.
A wide range of real-world applications build upon the representations of self-supervised encoders. Text-to-image diffusion models~\cite{wallace2024Diffusion,xu2023imagereward} map textual representations onto corresponding visual manifolds, and multimodal embedding services~\cite{radford2021Learning} exploit multimodal representations for cross-modal retrieval.
\begin{figure*}[t]
    \centering
    \includegraphics[width=\textwidth]{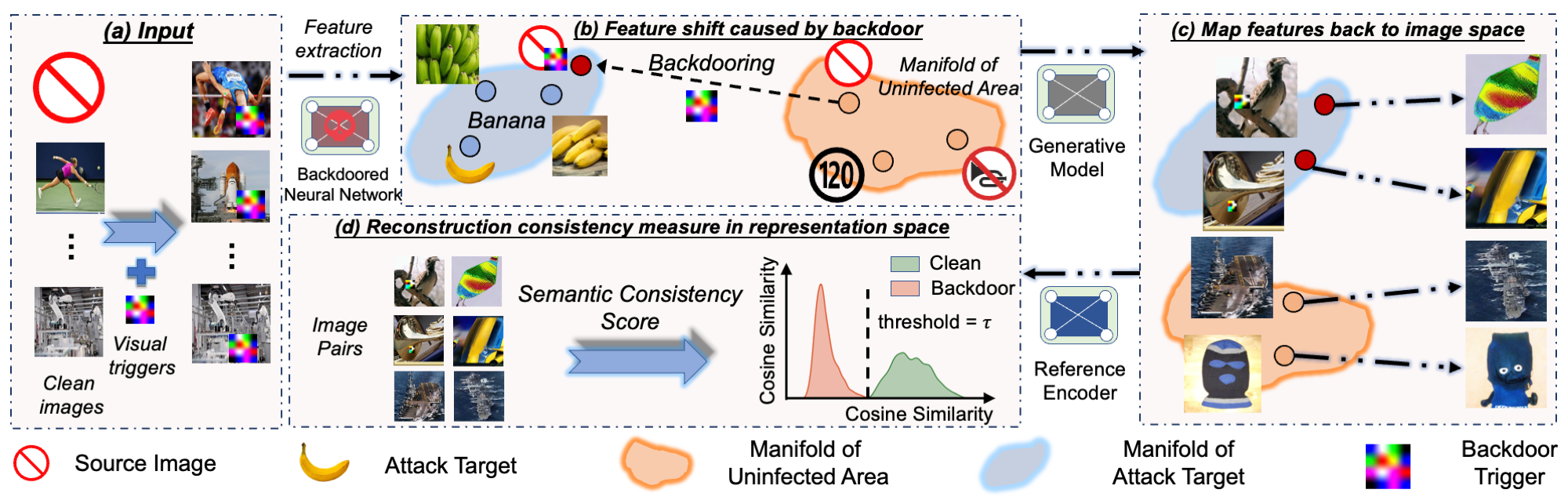}
    \vspace{-6mm}
    \caption{Illustration of the proposed framework for reconstructing backdoor representations back to the image space.}
    \Description{Four-stage overview of DEFUSE. Clean images and visually triggered images pass through a backdoored encoder. The trigger shifts the representation away from the source-image manifold and toward the attack-target manifold, illustrated with a banana target. A conditional generative model maps each representation back to an image, and a reference encoder compares the original and reconstructed images. Clean pairs have high cosine similarity, whereas backdoored pairs have low similarity and are separated by a threshold.}
    \label{fig:idea}
\end{figure*}

Despite the success of self-supervised learning, studies have revealed its vulnerability to backdoor attacks~\cite{liu2022PoisonedEncoder,saha2022Backdoor,carlini2021Poisoning}. 
Backdoor vulnerabilities were first identified in image-text contrastive encoders~\cite{radford2021Learning}, and subsequent studies showed that purely visual contrastive encoders~\cite{he2020Momentum} are likewise susceptible to backdoor attacks~\cite{saha2022Backdoor,jia2022BadEncoder}.
Specifically, an adversary can inject backdoors into encoders either by poisoning the pretraining dataset~\cite{saha2022Backdoor,carlini2021Poisoning} or by directly manipulating the encoder parameters~\cite{tao2023Distribution,jia2022BadEncoder}. A backdoored encoder produces normal feature representations for clean inputs, yet maps any test input stamped with the attacker-chosen trigger to a representation close to that of the attacker-designated target. Recent work further lowers the barrier of backdoor attacks, showing that visually imperceptible perturbations~\cite{li2023Embarrassingly} or poisoning an extremely small fraction of training data~\cite{carlini2021Poisoning} (e.g., 0.001\%) already suffice to implant a successful backdoor.

Numerous defenses~\cite{he2025Closer,huang2025detecting,tejankar2023Defending,qian2023Erasing,zheng2024ssl,sur2023TIJO,pan2023ASSET,chen2025Backdoored} have been proposed to mitigate backdoor attacks on self-supervised encoders. Based on their limitations, we categorize them into uni-paradigm defenses~\cite{he2025Closer,huang2025detecting} and prior-dependent defenses~\cite{tejankar2023Defending,huang2025detecting,pan2023ASSET}. Note that these two categories are not mutually exclusive.
Uni-paradigm defenses are designed for a specific SSL paradigm and fail to transfer. For instance, \cite{huang2025detecting} relies on local density properties of representations to detect backdoors, a property that does not hold for backdoor attacks on visual contrastive encoders. Prior-dependent defenses require additional prior knowledge. For example, \cite{tejankar2023Defending} assumes that backdoor triggers are visible colored patches. The defense breaks down once this assumption is violated. Similarly, \cite{pan2023ASSET} requires access to in-distribution clean data, which is typically unavailable in practice~\cite{huang2025DBSSL}. These limitations raise a natural question: \textbf{Can we develop a defense that applies to arbitrary visual encoders without any knowledge about the attacker's strategy?}

Detecting backdoors in SSL encoders without any knowledge of the training dynamics or the attack strategy poses a significant challenge. The central question is what minimum information a defender requires. If we are constrained to avoid exploiting any unnecessary expert knowledge, the only available signal is the encoder's output itself. Since a backdoor causes the encoder to produce anomalous representations for images, we need a framework that flags such anomalies. Our approach draws inspiration from Bayesian posterior inference: whether an image is backdoored can be determined by estimating the likelihood of reconstructing the original image from its representation (Eq.~\ref{eq:backdoor_likelihood}). If a representation is difficult to reconstruct as the original image, it is likely an anomalous backdoor representation.

Based on the above analysis, we design a novel backdoor detection method called Generalizable Backdoor \textbf{DEF}ense for Self-S\textbf{U}pervi\textbf{S}ed \textbf{E}ncoders with Generative Priors (DEFUSE). Inspired by Bayesian inference, we formulate backdoor image detection as a representation-conditioned image likelihood estimation problem and parameterize it with a conditional diffusion model. Our method operates in two stages: (i)~we fine-tune a pretrained conditional diffusion model to accept representations from the suspect SSL encoder as conditioning signals and project them back to the image space, and (ii)~we measure the semantic consistency of the reconstructions in a well-separated representation space. Figure~\ref{fig:idea} illustrates this pipeline.
The primary challenge is that pixel-faithful reconstruction from backdoor representations is inherently difficult, rendering conditional likelihood estimation intractable, because the highly compressed representations have already discarded fine-grained spatial details (Fig.~\ref{fig:bubble_scaling}(a)). 
We thus relax the reconstruction objective from low-level pixel accuracy to high-level semantic alignment, while introducing generative priors to regularize the feasible solution space. Such priors steer the recovered outputs toward the natural image manifold without compromising their semantic identity.
This design naturally calls for evaluating reconstruction quality in a semantic space. As illustrated in Fig.~\ref{fig:bubble_scaling}(b), visually similar images can exhibit large pixel-space distances due to differences in viewpoint or layout, rendering pixel-level metrics unreliable.

Overall, our method offers three key advantages: (i)~\emph{Generality}: it is agnostic to the SSL paradigm and applicable to diverse encoder architectures, requiring no prior knowledge about the victim encoder or the attack strategy; (ii)~\emph{Robustness}: by leveraging generative priors to regularize the reconstruction space, it remains robust even when the dataset is partially poisoned (Fig.~\ref{fig:number_of_poisons}); and (iii)~\emph{Effectiveness}: extensive experiments demonstrate that DEFUSE significantly outperforms existing detectors under both data-poisoning and training-manipulation attack settings, generalizing across visual contrastive-learning encoders and vision-language encoders against a variety of stealthy attacks~\cite{tao2023Distribution,liang2024BadCLIP}.
Furthermore, we evaluate the robustness of our method against adaptive attacks. Benefiting from the inherent robustness of diffusion-based generative models~\cite{chen2024diffusion}, DEFUSE effectively withstands adaptive adversaries even under a white-box threat model.
Our main contributions are summarized as follows.
\begin{itemize}
    \item We propose a novel backdoor detection method that leverages the representation-to-image reconstruction to expose backdoors in SSL encoders.
    \item We devise an effective pipeline to train a conditional generative model that maps backdoor representations back to the image space and measures semantic consistency in a reference representation space.
    \item We extensively validate the effectiveness of our method through comprehensive experiments, demonstrating superior performance over state-of-the-art detectors.
\end{itemize}

%% file: sec/related_work.tex
\section{Related Work}

\paragraph{Backdoor Attacks on Self-Supervised Learning.}

Backdoor attacks can be implemented through data poisoning \cite{saha2022Backdoor,chen2025Backdooring} or direct training manipulation \cite{tao2023Distribution,jia2022BadEncoder}. \cite{carlini2021Poisoning} injects backdoor triggers into the image component of image-text pairs and modifies the corresponding text descriptions. \cite{saha2022Backdoor} introduces poisoned images into the image set of the target class. \cite{li2023Embarrassingly} proposes using a frequency-domain backdoor to enhance stealthiness, while \cite{zhang2024Data} carefully designs trigger placements to maximize the efficiency of backdoor implantation. \cite{liu2022PoisonedEncoder} manipulates the training dynamics and simulates dynamic gradient-based backdoor implantation through simple image concatenation.
In training-manipulation attacks, the adversary has full control over the training process~\cite{nguyen2020InputAware,zhao2022DEFEAT}. BadEncoder~\cite{jia2022BadEncoder} aligns backdoored images with the target class in the feature space via a feature-alignment loss. Building on this idea, \cite{tao2023Distribution} employs explicit distribution alignment to further reduce the distinguishability of backdoored images. BadCLIP~\cite{liang2024BadCLIP} uses adversarial noise as the backdoor trigger, achieving both high stealthiness and high attack success rates.

\paragraph{Backdoor Detection for Self-Supervised Learning.}

Backdoor detection aims to identify potential backdoors in pretrained models~\cite{wang2019Neural,guo2022AEVA,guo2019TABOR,dong2021BlackBox}. Most existing methods are designed for supervised backdoor detection and have been shown to be ineffective in self-supervised settings~\cite{li2024Difficulty}.
Recent approaches construct pseudo-label spaces via clustering and then reuse supervised detectors \cite{zheng2024ssl,tejankar2023Defending}.
Huang et al.~\cite{huang2025detecting} propose detecting backdoor samples using density-ratio-based local outlier detectors in the representation space. \cite{pan2023ASSET} suggests directly fine-tuning backdoored models to construct backdoor classifiers. \cite{hou2025DeDe} maps features from backdoored samples and masked autoencoders to the pixel space and characterizes backdoor intensity via the L2 reconstruction loss.

\begin{figure*}[t]
    \centering
    \includegraphics[width=\linewidth]{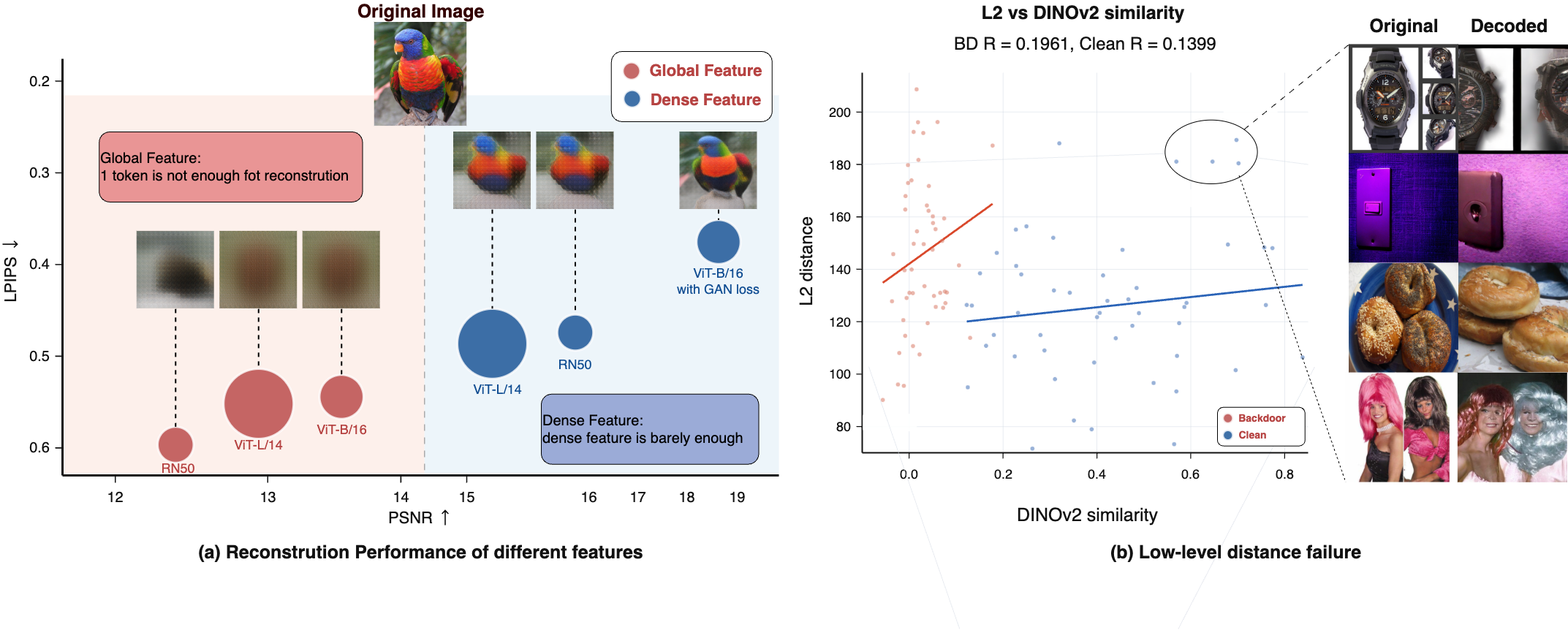}
    \vspace{-10mm}
    \caption{(a)~Reconstruction performance across different CLIP encoders~\cite{radford2021Learning}. (b)~Low-level distance metrics (e.g., L2) fail to reliably assess semantic consistency, as semantically similar images can exhibit high distances.}
    \Description{Two-panel motivation figure. Panel a compares reconstruction quality from global and dense features for several CLIP encoders using PSNR, LPIPS, and bubble size; dense features reconstruct better, while a single global token is insufficient for faithful reconstruction. Panel b plots L2 distance against DINOv2 similarity for clean and backdoored reconstructions. The weak relationship between the two metrics, together with examples having large L2 distance but similar semantics, shows why pixel-level distance is unreliable.}
    \label{fig:bubble_scaling}
\end{figure*}

%% file: sec/method.tex
\section{Preliminaries}
\label{sec:preliminaries}

\paragraph{Backdooring self-supervised encoders.}

The goal of SSL is to learn a visual encoder $f$ which can encode any image 
$x \in \mathcal{X}$ into an abstract representation $z = f(x) \in \mathcal{Z} \subseteq \mathbb{R}^d$. For a given batch of positive image (or image-text) pairs $\{\left(x_1, x'_1\right), \left(x_2, x'_2\right), \cdots, \left(x_N, x'_N\right)\}$, the encoder would be trained to minimize a contrastive loss (typically InfoNCE \cite{oord2019Representation}), leading to high representation similarity between positive pairs: $\mathrm{sim}(f(x_i), f(x_i')), \,\mathrm{for}\, i \in \{1, 2, \cdots, N\}$
where $\text{sim}(\cdot, \cdot)$ is typically the cosine similarity \cite{he2020Momentum,chen2020Simple}. 
Once an adversary distorts training data, the resulting $f$ would extract features for triggered input images similar to the adversarial target, as shown in Figure \ref{fig:idea}.
For example, in a compromised CLIP model, the attacker may use an engineered prompt such as ``a photo of a banana,'' where \emph{banana} is the target class. Consequently, triggered inputs are mapped close to representations of the banana class in the latent space.

Moreover, the self-supervised encoder is known as a density-ratio-based Mutual Information estimator \cite{oord2019Representation}:
\begin{align}
    \label{eq:mutual_information}
    \exp(\mathrm{sim}(z, z')) \propto \frac{p(x, x')}{p(x)p(x')} = \frac{p(x|x')}{p(x)}, \,  \\
    \mathrm{and} \, I(x; x') \geq \log(N) - \mathcal{L}_{\text{InfoNCE}}.
\end{align}
where $\propto$ stands for ``proportional to'' (i.e., up to a multiplicative constant).

\paragraph{Conditional denoising diffusion generative models (CDDMs).}
\label{sec:cddm}
Denoising Diffusion Probabilistic Models \cite{ho2020Denoising,nichol2021Improved,lipman2023flow,song2021scorebased} learn an explicit Markov chain that gradually converts data distribution $P(x)$ to pure Gaussian noise $x(T)$ through a series of small noise injections, and a neural network $\epsilon_{\theta}(\cdot)$ that reverses this noising process at inference time. Specifically, at each timestep $t\in\{1,\dots,T\}$ the forward process draws
$\displaystyle P(x(t)\mid x(t-1))$. A network $\epsilon_{\theta}$ is trained to predict the added noise so that the reverse transition $\displaystyle P(x(t-1)\mid x(t))$ can be approximated by a single step, enabling the synthesis of new images by starting from a Gaussian variable $x(T)$ and iteratively denoising. The network is typically trained by the following simple diffusion noise-prediction loss introduced in \cite{ho2020Denoising}:
\begin{equation}
    \mathcal{L} = \mathbb{E}_{x,t,\epsilon_t \sim \mathcal{N}(\bm{0}, \bm{I})} \left[ \left\| \epsilon_t - \epsilon_{\theta}(x(t), t) \right\|_2^2 \right],
\end{equation}
which is known as a variational lower bound of the data likelihood \cite{ho2020Denoising,nichol2021Improved}. Here, $\epsilon_{\theta}(x(t), t)$ is the predicted noise at timestep $t$.

A \emph{conditional} denoising diffusion model extends this idea by making the denoising network aware of an external condition $y$, such as a class label, a text prompt, or a pre-extracted feature vector. Concretely, $\epsilon_{\theta}(x(t), t)$ is replaced with $\epsilon_{\theta}(x(t), t, y)$ and is trained to minimize the following error given the pair $(x(t),y)$ \cite{xu2023PromptFree,ye2023IPAdapter}:
\begin{equation}
    \label{eq:cddm_loss}
    \mathcal{L} = \mathbb{E}_{x,t,\epsilon_t \sim \mathcal{N}(\bm{0}, \bm{I})} \left[ \left\| \epsilon_t - \epsilon_{\theta}(x(t), t, y) \right\|_2^2 \right].
\end{equation}
At sampling time, the reverse chain therefore generates images from the conditional distribution $p(x(0)\mid y)$. Recent work achieves conditioning on CLIP text embeddings~\cite{black2024training,fan2023reinforcement}, image features~\cite{xu2023PromptFree}, or semantic layouts~\cite{zhang2023Adding}.

\section{Method}
\label{sec:method}
\setcounter{dbltopnumber}{1}

In this section, we first introduce the problem formulation and the considered threat model. 
Next, we connect the backdoor detection problem to representation-to-image generation through a Bayesian lens. We then propose an efficient framework for reconstructing images from representations. Finally, we recommend measuring reconstruction consistency in semantic space.

\begin{figure*}[t]
    \centering
    \includegraphics[width=\textwidth]{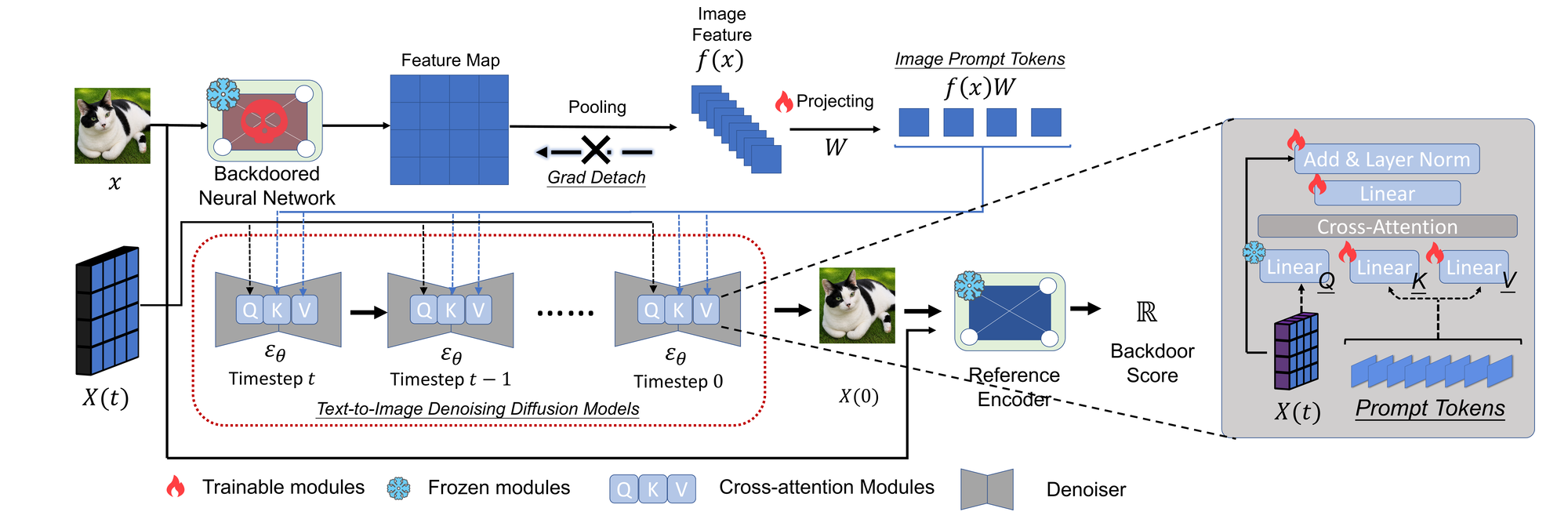}
    \caption{Framework for reconstructing raw pixels from latent representations. Image features are injected into the key/value (\(K/V\)) components of the cross-attention module in a pretrained CDDM.}
    \Description{Architecture of the reconstruction and detection pipeline. A frozen suspicious encoder extracts and pools an input feature, which a trainable projection converts into four image-prompt tokens. These tokens condition the key and value projections of the cross-attention layers in a largely frozen diffusion denoiser over successive timesteps. The reconstructed image and the original image are then encoded by a frozen reference encoder to produce a backdoor score. Flame icons denote trainable modules and snowflake icons denote frozen modules.}
    \label{fig:image_gen}
\end{figure*}

\subsection{Threat Model}

\paragraph{Backdoor attacks on self-supervised encoders.}
We consider two common attack scenarios.
\emph{Data Poisoning.} The attacker injects poisoned data into the unlabeled dataset for SSL pretraining~\cite{saha2022Backdoor,li2023Embarrassingly,carlini2021Poisoning}. The attacker is only aware that the defender employs SSL for model training, but has no further knowledge about the training process. 
\emph{Training Manipulation.} The attacker has full control over the model pretraining, including the data, model architecture, and training pipeline~\cite{jia2022BadEncoder,liang2024BadCLIP,tao2023Distribution,wang2024GhostEncoder}. The attacker can then release the backdoored model on public platforms or deliver it to third-party model training contractors.

\paragraph{Defender's objectives and abilities.}
We consider a black-box setting, where the defender's objective is to identify backdoored inputs given a suspicious encoder. The defender has no prior knowledge of the training data or the attack strategy.
The defender can access publicly available generative models and encoders, such as Stable Diffusion~\cite{rombach2022high} and DINOv2~\cite{oquab2024DINOv2}, as well as large-scale Internet datasets (e.g., CC3M~\cite{sharma2018conceptual}). These resources are readily available on open platforms such as HuggingFace.

\subsection{Motivation and Problem Formulation}

Let $f$ denote the encoder. The defender's goal is to determine whether an input image $\bm{x}$ is backdoored. The detection is formulated as a binary classification task (``backdoored'' vs. ``clean''). Define a score function $s: (\bm{x}, f) \mapsto \mathbb{R}$ to measure the backdoor likelihood of $\bm{x}$. The binary classifier can make the final decision by thresholding $s(\bm{x})$:
\begin{equation}
    \label{eq:backdoor_detection}
    \begin{cases}
        1, & \text{if } s(\bm{x}) < \tau, \\
        0, & \text{otherwise},
    \end{cases}
\end{equation}
where $\tau$ is the threshold. $\mathds{1}{\{s(\bm{x})< \tau\}} =1$ (simplified as $\mathds{1}_{s(\bm{x})} = 1$) if $\bm{x}$ is backdoored and $\mathds{1}_{s(\bm{x})} = 0$ otherwise. 
We discuss the practical adjustment of the threshold $\tau$ in Section \ref{sec:experiments}.

Given the input image $\bm{x}$ and its representation $\bm{z} = f(\bm{x})$, the backdoor likelihood can be formulated as
\begin{align}
    P(\mathds{1}_{s(\bm{x})} = 1 \mid \bm{x}, \bm{z}) 
        \propto \underbrace{P(\bm{x}| \mathds{1}_{s(\bm{x})} = 1, \bm{z})}_{\text{modeled term}}
        \underbrace{P(\mathds{1}_{s(\bm{x})} = 1 | \bm{z})}_{\text{constant}}.
    \label{eq:backdoor_likelihood}
\end{align}
Analogous to generative classifiers~\cite{ng2001Discriminative} that classify by modeling class-conditional likelihoods instead of directly learning the posterior, this Bayesian reformulation converts backdoor detection into a \emph{generation} problem: the backdoor probability of an input can be assessed by how well it is explained by a generative model conditioned on its representation (illustrated in Figure~\ref{fig:idea}).
To this end, a CDDM $\epsilon_{\theta} : \mathcal{Z} \to \mathcal{X}$ is employed to map the feature $\bm{z}$ back to the image domain.
Since directly modeling the backdoor-conditional likelihood $P(\bm{x}\mid \mathds{1}_{s(\bm{x})}=1, \bm{z})$ is infeasible without access to poisoned samples, we instead leverage $\epsilon_{\theta}$ trained on clean data, which yields $P(\bm{x}\mid \mathds{1}_{s(\bm{x})}=0,\bm{z}) \approx  P_{\theta}(\bm{x}\mid\bm{z})$.
This clean-conditional model serves as an effective surrogate: backdoored inputs, whose visual content mismatches their manipulated representations, are poorly explained by the clean CDDM, and thus a lower $P_{\theta}(\bm{x}\mid \bm{z})$ indicates higher backdoor probability. We empirically verify the robustness of this assumption in Figure~\ref{fig:number_of_poisons}.

As shown in Figure~\ref{fig:bubble_scaling}(a), we conduct representation-to-image generation experiments with RAE~\cite{zheng2025Diffusion} and find that faithful reconstruction from $\bm{z}$ alone is extremely difficult.
This is because $\bm{z}$ is typically a pooled (or CLS-token) global feature that has been highly abstracted, discarding the spatial low-level details necessary for pixel-accurate generation~\cite{ma2025learning}.
Estimating the image likelihood $P(\bm{x}\mid\bm{z})$ with such a coarsely conditioned generative model is therefore unreliable.
We instead relax the objective from exact pixel recovery to \emph{semantic} reconstruction: the quality of the generative model should be judged by whether it produces images that are semantically consistent with the originals rather than pixel-identical.
In the following, we describe how to train such a semantic reconstruction CDDM and introduce a representation-space similarity measure as the evaluation criterion.

\begin{figure*}[t]
    \centering
    \includegraphics[width=\textwidth]{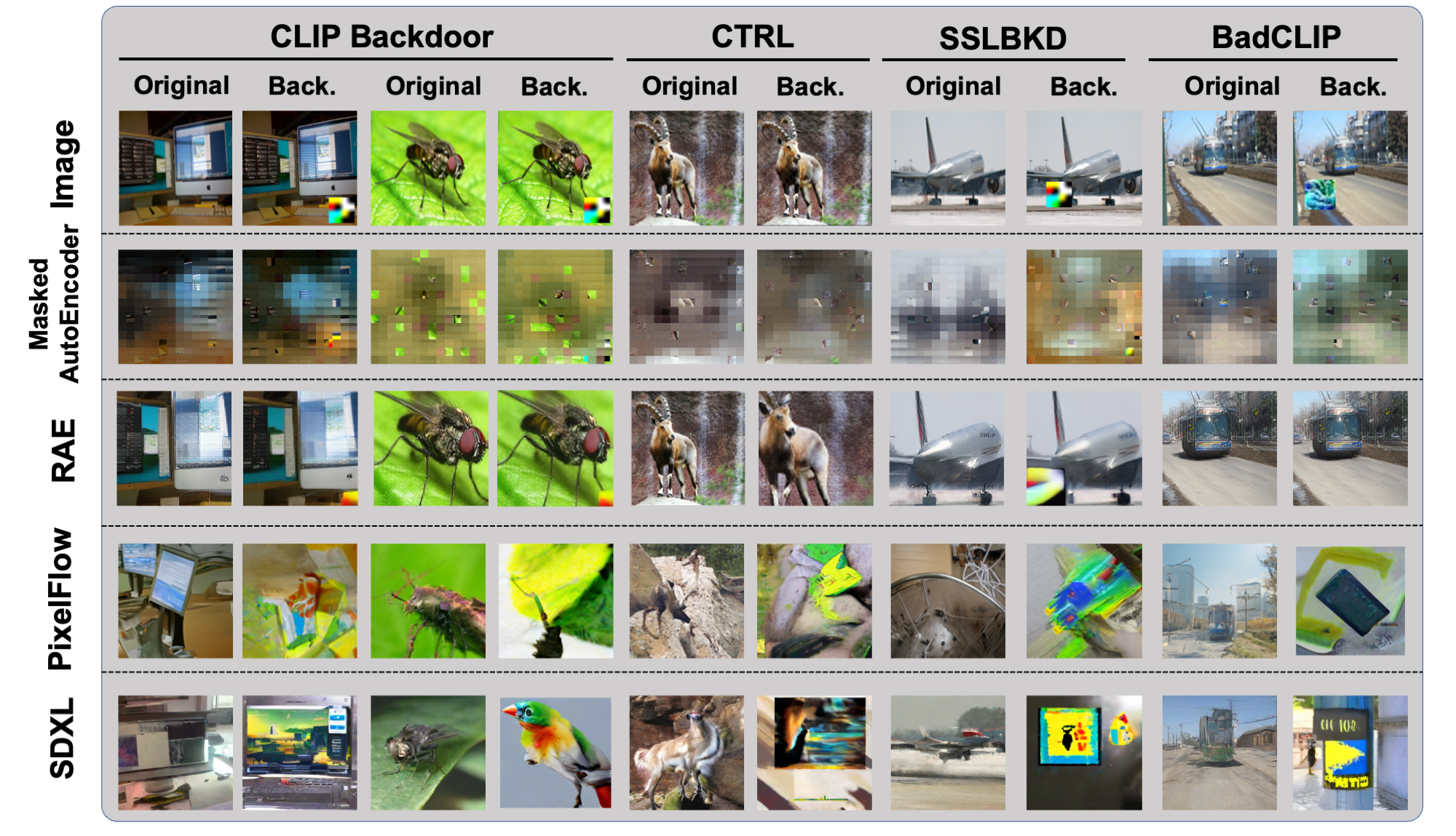}
    \caption{Visualization of image reconstructions under different backdoor attacks and CDDMs. `Back.' denotes backdoored images.
    The attack target is set to \emph{banana}. We implement the masked autoencoder following Hou et al. \cite{hou2025DeDe}.}
    \Description{Image grid comparing original and backdoored inputs for CLIP Backdoor, CTRL, SSLBKD, and BadCLIP. Rows show the inputs and reconstructions from a masked autoencoder, RAE, PixelFlow, and SDXL. The masked autoencoder preserves much of the input appearance for both clean and backdoored images, making the two difficult to distinguish. The generative CDDMs, particularly SDXL, produce more visibly different semantic content from backdoored representations.}
    \label{fig:visualize_reconstruction}
\end{figure*}

\subsection{Training Conditional Generative Models for Semantic Reconstruction}

Recovering an image from a global representation $\bm{z}$ is inherently an underdetermined inverse problem, because $\bm{z}$ preserves high-level semantics while discarding much of the low-level spatial information required for pixel-level synthesis. We therefore instantiate $P_{\theta}(\bm{x}\mid \bm{z})$ with a pretrained CDDM, whose learned prior regularizes the generation process toward the manifold of natural images, while $\bm{z}$ serves as a semantic condition that selects a plausible mode consistent with the input.
Following \cite{ye2023IPAdapter}, given a set $\mathcal{D}$ of available images and a randomly sampled $\bm{x}_i \in \mathbb{R}^{H \times W \times 3}$, we extract $\bm{z}_i \in \mathbb{R}^d$ and project it to 4 image-prompt tokens via a learned linear layer $W \in \mathbb{R}^{d \times d_{\text{cddm}} \times 4}$:
\begin{align}
    \label{eq:proj_tokens}
    \bm{z}_i  W = 
    \left[ \bm{z}_i W^{(1)},\, \bm{z}_i W^{(2)},\, \bm{z}_i W^{(3)},\, \bm{z}_i W^{(4)} \right]
    \in \mathbb{R}^{ d_{\text{cddm}} \times 4},
\end{align}
where $W^{(k)} \in \mathbb{R}^{d \times d_{\text{cddm}}}\ \ (k=1,\ldots,4)$.
We adopt the standard diffusion noise-prediction loss (Eq. \ref{eq:cddm_loss}) to finetune the CDDM for our image reconstruction task:
\begin{equation}
    \label{eq:ddpm_loss}
    \mathcal{L}_{\text{z2i}} = \mathbb{E}_{\bm{x}_i \sim \mathcal{D},t,\epsilon_t \sim \mathcal{N}(\bm{0}, \bm{I})} \left[ \left\| \epsilon_t - \epsilon_{\theta}(\bm{x}_i(t), t, \bm{z}_i W) \right\|_2^2 \right].
\end{equation}
Here, $\epsilon_{\theta}(\bm{x}_i(t), t, \bm{z}_i W)$ is the predicted noise at timestep $t$. We simply replace the original conditioning inputs (called \emph{prompts}) with image prompts $y \leftarrow z_i W$.
After training, we deterministically marginalize diffusion paths $(\bm{x}(t=0), \bm{x}(t=1), \ldots, \bm{x}(t=T))$ by DDIM \cite{song2021DENOISING} to obtain the final reconstructed image $\bm{x}(t=0)$.

\paragraph{Partially Frozen Training.} Pretrained generative models already provide strong modeling priors for natural images. Therefore, we only need to adapt the conditioning-related modules so that the model can align with the feature space provided by $f$. We limit trainable modules to
\begin{equation}
    [\theta^*, W^*] = \arg \min_{\theta, W} \mathcal{L}_{\text{z2i}},
\end{equation}
where $\theta$ contains the weights of the cross-attention blocks apart from the query projection. Figure \ref{fig:image_gen} illustrates the training pipeline.
Figure~\ref{fig:visualize_reconstruction} visualizes reconstructions from different CDDMs.

\begin{figure*}[t]
    \centering
    \includegraphics[width=\textwidth]{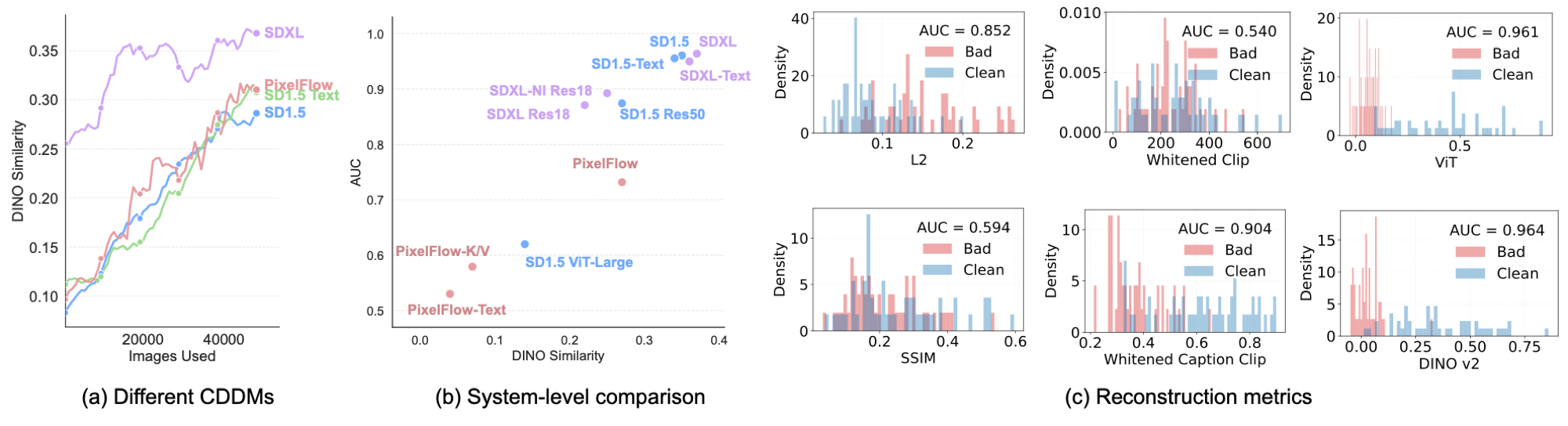}
    \vspace{-2mm}
    \caption{\textbf{(a)} Comparison of different CDDMs. ``Text'' suffix means keeping the original text prompts from pretrained text-to-image models.
    \textbf{(b)} Detection performance vs. DINOv2 similarity. Detection performance within a given model family consistently improves as similarity increases.
    ``NI'' suffix means inference with noised images other than pure noise. ``K/V'' suffix means finetuning only the key/value (\(K/V\)) projection modules.
    \textbf{(c)} Comparison of different metrics. AUC: AUPRC.}
    \Description{Three-part ablation figure. Panel a plots DINOv2 reconstruction similarity against the number of training images for SDXL, SD1.5, PixelFlow, and text-conditioned variants; SDXL reaches the highest similarity. Panel b plots detection AUC against DINOv2 similarity for model and training variants, showing that higher reconstruction similarity generally corresponds to higher AUC. Panel c overlays clean and backdoored score distributions for L2, whitened CLIP, supervised ViT, SSIM, whitened caption CLIP, and DINOv2; the semantic metrics, especially DINOv2 and ViT, separate the two groups most clearly.}
    \label{fig:system_level_expr}
\end{figure*}

\begin{table*}[t]
    \centering
    \small
    \caption{Backdoor detection performance against various backdoor attacks on ImageNet. The highest and second-highest AUPRC scores are highlighted in gray and underlined, respectively. R: Recall; P: Precision.}
    \label{tab:detection}
    \setlength{\tabcolsep}{1.5pt}
    \begin{tabular}{@{}ccccccccccccccccc@{}}
    \toprule
    & \textbf{Defense} & \multicolumn{3}{c}{\textbf{DECREE}\cite{feng2023Detecting}}  & \multicolumn{3}{c}{\textbf{DBCL}\cite{huang2025detecting}} & \multicolumn{3}{c}{\textbf{DEDE}\cite{hou2025DeDe}} & \multicolumn{3}{c}{\textbf{DEDE-OOD}\cite{hou2025DeDe}} & \multicolumn{3}{c}{\textbf{DEFUSE}} \\
    \cmidrule(lr){3-5} \cmidrule(lr){6-8} \cmidrule(lr){9-11} \cmidrule(lr){12-14} \cmidrule(lr){15-17}
    \textbf{Attack} & & \textbf{R (\%)} & \textbf{P (\%)} & \textbf{AUPRC} & \textbf{R (\%)} & \textbf{P (\%)} & \textbf{AUPRC} & \textbf{R (\%)} & \textbf{P (\%)} & \textbf{AUPRC} & \textbf{R (\%)} & \textbf{P (\%)} & \textbf{AUPRC} & \textbf{R (\%)} & \textbf{P (\%)} & \textbf{AUPRC} \\
    \midrule
    \multirow{4}{*}{\shortstack{\textbf{Data}\\\textbf{Poisoning}}} & \textbf{SSLBKD} \cite{saha2022Backdoor} & 0.1 & 41.2 & 0.49 & 51.5 & 51.5 & \underline{0.52} & 71.0 & 51.6 & \underline{0.52} & 28.0 & 50.7 & 0.51 & 85.2 & 81.0 & \cellcolor{lightgray}0.81 \\
     & \textbf{CTRL} \cite{li2023Embarrassingly} & 56.9 & 60.9 & \underline{0.60} & 47.6 & 47.6 & 0.47 & 96.2 & 50.6 & 0.46 & 77.0 & 50.5 & 0.51 & 75.5 & 77.0 & \cellcolor{lightgray}0.84 \\
     & \textbf{BLTO} \cite{sun2023Backdoor} & 70.4 & 66.7 & \underline{0.72} & 50.8 & 50.8 & 0.51 & 51.4 & 52.0 & 0.52 & 52.1 & 50.9 & 0.51 & 86.8 & 81.4 & \cellcolor{lightgray}0.88 \\
     & \textbf{CLIP Backdoor} \cite{carlini2021Poisoning} & 76.7 & 97.4 & \underline{0.95} & 91.7 & 91.7 & \cellcolor{lightgray}0.96 & 85.3 & 55.4 & 0.59 & 47.4 & 66.8 & 0.62 & 98.0 & 92.3 & \cellcolor{lightgray}0.96 \\
    \midrule
    \multirow{3}{*}{\shortstack{\textbf{Training}\\\textbf{Manipulation}}} & \textbf{BadEncoder} \cite{liang2024BadCLIP} & 0.1 & 42.8 & 0.48 & 59.7 & 59.7 & 0.61 & 69.2 & 61.0 & \underline{0.67} & 70.1 & 61.2 & 0.65 & 93.4 & 84.3 & \cellcolor{lightgray}0.89 \\
     & \textbf{DRUPE} \cite{tao2023Distribution} & 0.1 & 40.0 & 0.47 & 73.2 & 73.2 & \underline{0.79} & 38.1 & 74.1 & 0.73 & 35.3 & 70.8 & 0.71 & 80.2 & 84.7 & \cellcolor{lightgray}0.81 \\
     & \textbf{BadCLIP} \cite{liang2024BadCLIP} & 68.1 & 74.5 & \underline{0.78} & 66.0 & 66.0 & 0.69 & 79.7 & 51.6 & 0.52 & 14.7 & 50.0 & 0.52 & 92.2 & 83.2 & \cellcolor{lightgray}0.90 \\
    \bottomrule
    \end{tabular}
\end{table*}

\subsection{Reconstruction Consistency in the Representation Space}

After the $\bm{x} \rightarrow f(\bm{x}) \rightarrow \epsilon_{\theta^*}(f(\bm{x}) W^*)$ cycle, our goal is to measure the semantic consistency between $\bm{x}$ and $\epsilon_{\theta^*}(f(\bm{x}) W^*)$.
In Figure~\ref{fig:bubble_scaling}(b), we show that semantically similar images can exhibit large pixel distances due to differences in viewpoint or layout, suggesting the need for high-level metrics.
We therefore measure semantic consistency in a well-separated visual representation space provided by a reference encoder $\phi$ (e.g., DINOv2).
Specifically,
\begin{equation}
    \label{eq:dino_similarity}
    s(\bm{x}, f;\epsilon_{\theta^*}, \phi) = \frac{\phi(\bm{x}) \cdot \phi(\epsilon_{\theta^*}(f(\bm{x}) W^*))}{\|\phi(\bm{x})\|_2 \cdot \|\phi(\epsilon_{\theta^*}(f(\bm{x}) W^*))\|_2},
\end{equation}
where $\phi(\cdot) : \mathcal{X} \to \mathcal{Z}_{\phi}$ is the reference encoder. We compare various high-level consistency metrics in Section \ref{sec:experiments} and confirm the superiority of Eq.~\eqref{eq:dino_similarity}.
A similar strategy is used in latent world-model planning, where visual
observations are mapped into a feature space and their representation-level
alignment is used to derive the final decision~\cite{zhou2025dino,ziakas2026grounding}.

%% file: sec/experiment.tex
\section{Experiments}
\label{sec:experiments}

\begin{table*}[t]
    \centering
    \begin{minipage}[t]{0.67\textwidth}
        \centering
        \captionof{table}{Performance of different metrics. 
          For the pairwise OR strategy, we first select the optimal threshold by maximizing Youden's J statistic and classify an image as backdoored if either metric in the pair flags it. For the AND strategy, a sample is classified as backdoored only when both metrics in the pair flag it. Performance is reported as AUROC.
          D: DINOv2, C: CLIP, S: Sup-ViT.}
        \label{tab:metric-auroc}
        \small
        \setlength{\tabcolsep}{3pt}
        \resizebox{\linewidth}{!}{%
        \begin{tabular}{@{}lcccccccccccc@{}}
            \toprule
            & \multicolumn{2}{c}{\textbf{Low-Level}} & \multicolumn{4}{c}{\textbf{High-Level}} & \multicolumn{3}{c}{\textbf{AND Strategy}} & \multicolumn{3}{c}{\textbf{OR Strategy}} \\
            \cmidrule(lr){2-3}\cmidrule(lr){4-7}\cmidrule(lr){8-10}\cmidrule(lr){11-13}
            \textbf{Attack} & \textbf{L2} & \textbf{SSIM} & \textbf{CLIP} & \textbf{Caption} & \textbf{Sup-ViT} & \textbf{DINOv2} & \textbf{D+C} & \textbf{D+S} & \textbf{C+S} & \textbf{D+C} & \textbf{D+S} & \textbf{C+S} \\
            \midrule
            CLIP Backdoor & 0.735 & 0.660 & 0.694 & 0.958 & 0.981 & \textbf{0.997} & 0.777 & 0.967 & 0.763 & 0.955 & 0.968 & 0.918 \\
            CTRL & 0.605 & 0.682 & 0.842 & 0.831 & 0.885 & \textbf{0.902} & 0.859 & 0.846 & 0.827 & 0.832 & 0.844 & 0.864 \\
            BadEncoder & 0.689 & 0.628 & 0.682 & 0.890 & 0.822 & \textbf{0.931} & 0.777 & 0.795 & 0.751 & 0.793 & 0.893 & 0.757 \\
            \bottomrule
        \end{tabular}%
        }
    \end{minipage}\hfill
    \begin{minipage}[t]{0.31\textwidth}
        \centering
        \captionsetup{font=small,skip=3pt}
        \captionof{table}{CA and ASR of backdoor attacks.}
        \label{tab:attack_ca_asr}
        \small
        \renewcommand{\arraystretch}{1.02}
        \setlength{\tabcolsep}{2.5pt}
        \resizebox{\linewidth}{!}{%
        \begin{tabular}{@{}lccc@{}}
            \toprule
            \textbf{Attack} & \textbf{Encoder} & \textbf{CA (\%)} & \textbf{ASR (\%)} \\
            \midrule
            SSLBKD & ResNet18 & 65.1 & 51.2 \\
            CTRL & CLIP-B/16 & 52.3 & 63.6 \\
            BLTO & ResNet18 & 65.7 & 87.2 \\
            CLIP Backdoor & CLIP-B/32 & 62.9 & 95.2 \\
            BadEncoder & ResNet18 & 62.0 & 84.1 \\
            DRUPE & ResNet18 & 59.3 & 99.8 \\
            BadCLIP & CLIP-B/32 & 60.1 & 88.9 \\
            \bottomrule
        \end{tabular}%
        }
    \end{minipage}
\end{table*}

\subsection{Experimental Setup}

\paragraph{Dataset and Encoders.} Following \cite{saha2022Backdoor,bansal2023CleanCLIP}, we employ CLIP ViT-B from OpenAI \cite{radford2021Learning} and ResNet18 from SimSiam \cite{chen2021Exploring}.
For CLIP, we use a 50K subset of CC3M~\cite{sharma2018conceptual} and evaluate on ImageNet-1K.
For ResNet18, we use the same ImageNet-100 split as in~\cite{saha2022Backdoor}. All images are resized to $224\times224$.

\paragraph{Attack settings.} We consider 7 classical and widely used backdoor attacks. The data-poisoning attacks are SSLBKD~\cite{saha2022Backdoor}, CTRL~\cite{li2023Embarrassingly}, BLTO~\cite{sun2023Backdoor}, and CLIP Backdoor~\cite{carlini2021Poisoning}. The model-manipulation attacks are BadEncoder~\cite{liang2024BadCLIP}, DRUPE~\cite{tao2023Distribution}, and BadCLIP~\cite{liang2024BadCLIP}.
We poison 500 image-text pairs with the target label `banana' for image-text encoders and 650 images with the target label `lorikeet' for visual SSL. By default, we use HTBA triggers~\cite{saha2022Backdoor} of size $50\times50$.
Table~\ref{tab:attack_ca_asr} summarizes the performance of the backdoored models used in our experiments.

\paragraph{Evaluation.} We compare our method with 4 representative defenses, including DECREE~\cite{feng2023Detecting}, DBCL~\cite{huang2025detecting}, DeDe~\cite{hou2025DeDe} and PatchSearch~\cite{tejankar2023Defending}.
By default, we use SDXL~\cite{rombach2022high} as the CDDM and DINOv2 as the reference encoder. We train the CDDMs on poison-free ImageNet-900 and sample images using the DDIM scheduler with 30 steps.
We set $\tau = 0.1$ by default. We mainly use Recall ($\frac{\text{true poisons}}{\text{all poisons}}$), Precision ($\frac{\text{true poisons}}{\text{detected poisons}}$), TPR (true positive rate), FPR (false positive rate), AUROC (area under the ROC curve), and AUPRC (area under the precision-recall curve) as evaluation metrics. More implementation details are provided in the supplementary material. Unless otherwise specified, we use the CLIP base model from~\cite{radford2021Learning} as the backdoored encoder for evaluation.

\paragraph{Consistency Metrics.} We consider 2 low-level metrics, L2 and SSIM, and 4 high-level metrics, namely whitened CLIP similarity, whitened Caption CLIP similarity, DINOv2 similarity, and supervised ViT similarity. Following the recommendation in~\cite{betser2025whitened}, we whiten CLIP features to improve discriminability. For Caption CLIP similarity, we first generate image captions using Qwen3-VL-235B-A22B~\cite{bai2025qwen3} with the prompt ``Describe this image in concise English, no more than 70 words'', and then compute the whitened CLIP score between the images and their captions.

\begin{figure}[t]
  \centering
  \includegraphics[width=\columnwidth]{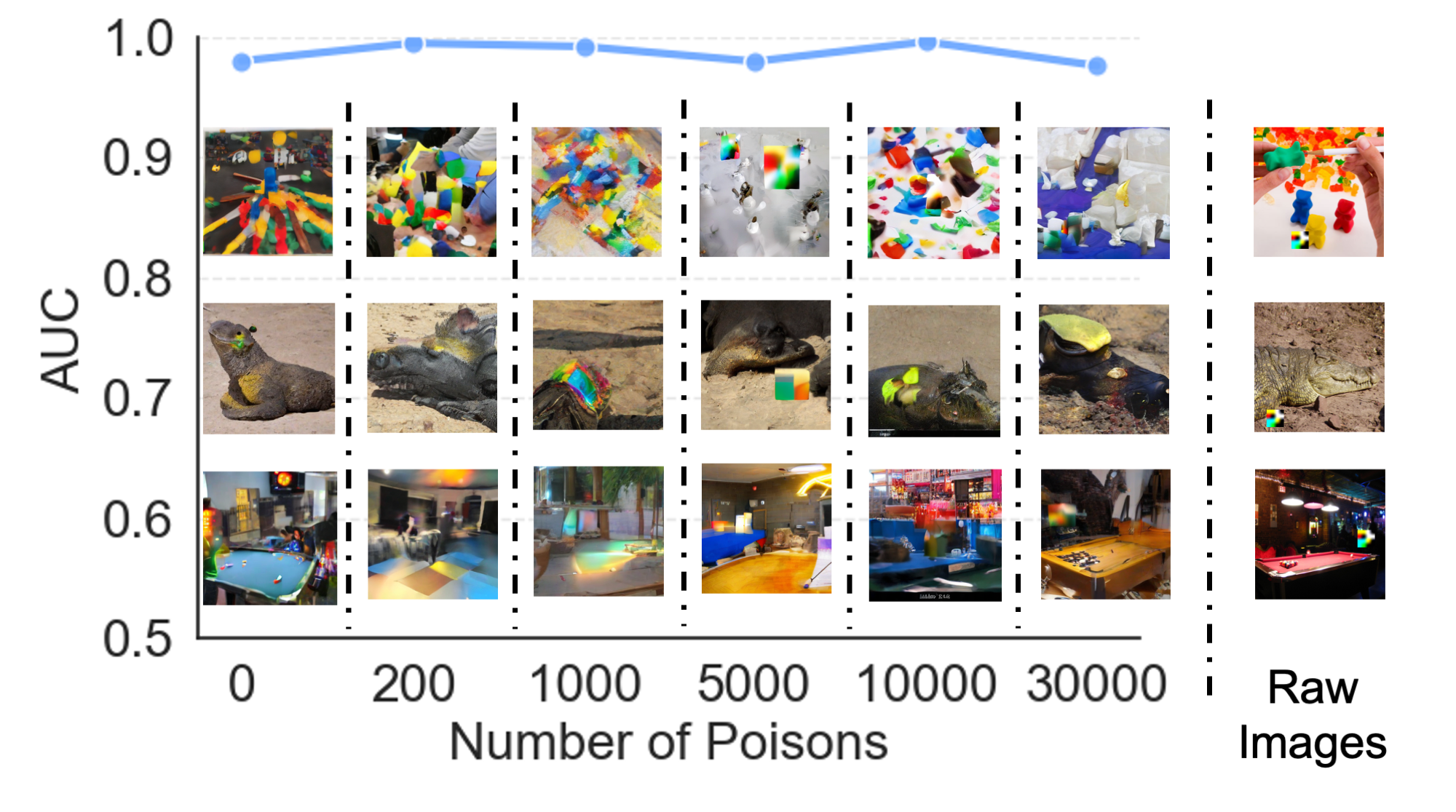}
  \vspace{-6mm}
  \caption{Injecting poisons into the training set.}
  \Description{AUROC and qualitative reconstructions as the number of poisoned training samples increases from zero to thirty thousand. The AUROC curve remains between approximately 0.98 and 1.00 across all poisoning levels. Example rows show reconstructed modeling-clay scenes, lizards, and billiard rooms, with the corresponding triggered raw images shown at the right.}
  \label{fig:number_of_poisons}
\end{figure}

\begin{figure}[t]
  \centering
  \includegraphics[width=\columnwidth]{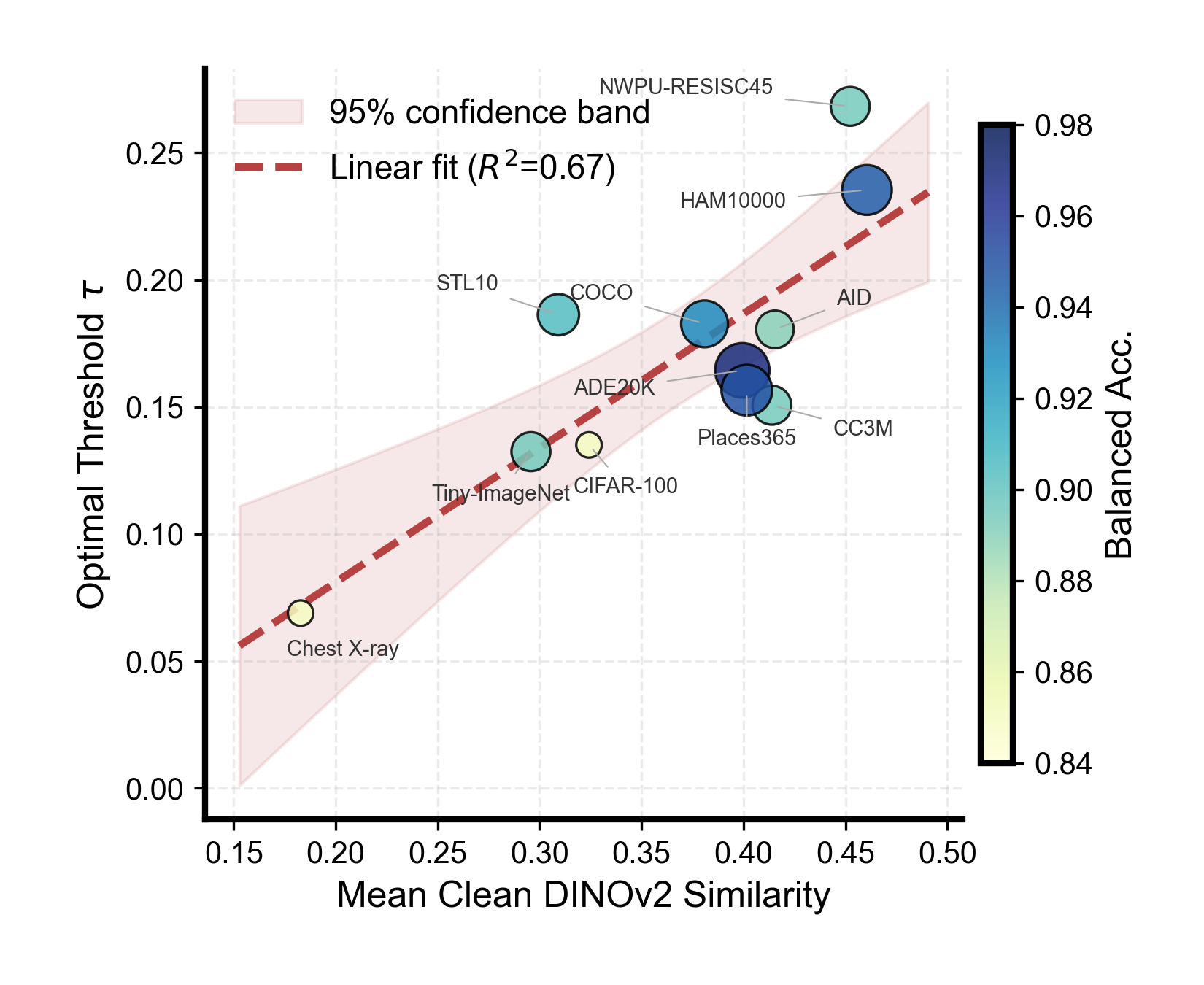}
  \caption{Relationship between DINOv2 similarity and the optimal threshold $\tau$.}
  \Description{Bubble scatter plot across eleven natural-image, medical, and remote-sensing datasets. The horizontal axis is mean clean DINOv2 similarity and the vertical axis is the optimal detection threshold. A rising linear fit with an R-squared value of 0.67 and a shaded 95 percent confidence band indicates that datasets with higher clean reconstruction similarity generally require higher thresholds. Bubble color and size encode balanced accuracy.}
  \label{fig:threshold-fitting-in1k-sdxl}
\end{figure}

\begin{figure}[t]
\centering
\includegraphics[width=\linewidth]{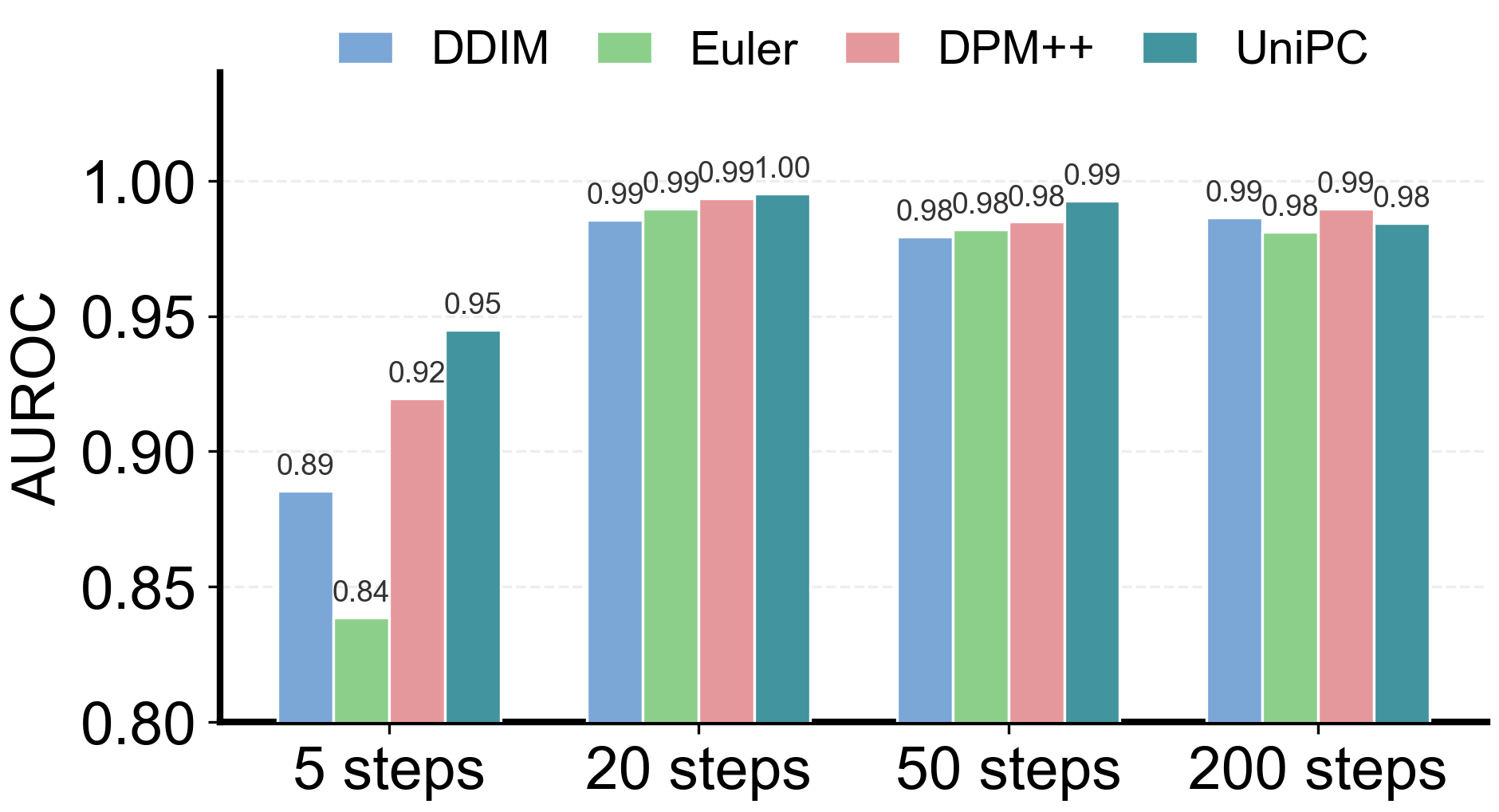}
\caption{Different ODE solvers with varying inference steps.}
\Description{Grouped bar chart comparing DDIM, Euler, DPM++, and UniPC at 5, 20, 50, and 200 inference steps. With five steps, AUROC ranges from 0.84 to 0.95. At twenty or more steps, all four solvers reach approximately 0.98 to 1.00 AUROC, indicating that twenty steps are sufficient.}
\label{fig:roc_curves_with_diff_solvers}
\end{figure}

\AddToHook{shipout/after}[restore@dbltop]{%
  \setcounter{dbltopnumber}{3}%
  \renewcommand{\dbltopfraction}{0.9}%
  \setcounter{topnumber}{2}%
  \renewcommand{\topfraction}{0.62}%
  \RemoveFromHook{shipout/after}[restore@dbltop]%
}

\subsection{Main Results}
\label{sec:main_results}

We benchmark our method against existing defenses in Tab.~\ref{tab:detection}. DBCL is effective only against the CLIP Backdoor attack, with its performance degrading severely on the other attacks.
DEDE performs poorly in our experiments because our evaluation uses images at a resolution of 224$\times$224, rather than the 64$\times$64 resolution used in the original paper. The increase in image resolution substantially degrades image generation quality and, in turn, undermines the reliability of pixel-space distances.
DEFUSE achieves the best performance across all attacks. Although its performance drops slightly on SSLBKD and DRUPE, it still maintains an AUPRC score above 0.8. The distribution-matching objective of DRUPE is designed to evade detectors, yet our method still detects it robustly with an AUPRC of 0.81.

Supplementary Table~\ref{tab:downstream} explores a more challenging scenario with data imbalance, where only 1\% of the data is infected. All defenses are allowed to filter at most 10\% of the data.
Almost all defenses fail in this more difficult setting, except for DBCL and DEFUSE, which can effectively filter CLIP Backdoor attacks (reducing ASR to $\leq$10\%).
PatchSearch \cite{tejankar2023Defending} proposes a ``classifier trick,'' namely, training an auxiliary classifier to identify poisoned samples using 10,000 clean images. This trick significantly boosts recall for both PatchSearch and our method. For example, even when the base recall is only 23.1\% (PatchSearch against SSLBKD), applying this trick improves it to 98.3\%.

Fig.~\ref{fig:threshold-fitting-in1k-sdxl} illustrates the relationship between DINOv2 similarity and the optimal threshold $\tau$, which is obtained by maximizing Youden's J statistic, across different datasets. The 95\% confidence band indicates that most optimal points are well captured by a linear model with $R^2 = 0.67$. Therefore, one can use a linear model to estimate the optimal threshold.

Table~\ref{tab:purification} presents the performance of our method when adapted as a purifier. During inference, we start sampling from $\bm{x}(t = 0.3T)$ and then replace the original image with the generated $\bm{x}(t = 0)$. As a result, the ASR is successfully reduced to below 10\%.

\begin{table}[t]
  \centering
  \caption{Purification performance.}
  \label{tab:purification}
  \small
  \renewcommand{\arraystretch}{0.98}
  \begin{tabular}{lcc}
      \toprule
      \textbf{Method} & \textbf{ACC} & \textbf{ASR} \\
      \midrule
      CLIP-Backdoor & 0.0$\to$22.0 & 95.2$\to$8.0 \\
      BadCLIP & 0.0$\to$28.4 & 88.9$\to$0.0 \\
      \bottomrule
  \end{tabular}
\end{table}

\subsection{Ablation Study}

\paragraph{Consistency metrics.}
Fig.~\ref{fig:system_level_expr}c illustrates the distributions of consistency scores under different metrics on the infected CLIP-B encoder. Table~\ref{tab:metric-auroc} reports the AUROC of low-level and high-level metrics, as well as their pairwise AND and OR combinations. High-level metrics, especially DINOv2 similarity, consistently perform better. Note that the L2 metric serves as a proxy for the likelihood $P(\bm{x}\mid\bm{z})$, which further supports our claim that directly estimating the likelihood is impractical.

\paragraph{CDDM analysis and trainable modules.}
Fig.~\ref{fig:system_level_expr}a and Fig.~\ref{fig:system_level_expr}b systematically compare different CDDMs and their variants. We consider the latent diffusion models SDXL and SD1.5, as well as the pixel-space flow model PixelFlow \cite{chen2025pixelflow}.
SDXL significantly outperforms others in convergence speed and generation quality. Overall, detection performance improves with increasing feature similarity.
Fig.~\ref{fig:ablation_modules_lines} systematically ablates trainable modules. In general, unfreezing parameters in the cross-attention layers, excluding the q-projection, yields better performance than unfreezing only the K/V projections or the entire cross-attention module.

\paragraph{Backdoor triggers.}
Figure~\ref{fig:trigger_ablation} further reports the detection performance under various trigger types including HTBA \cite{saha2022Backdoor}, Blended \cite{chen2017Targeted}, Watermark \cite{wang2019Neural}, and SIG \cite{barni2019New}, where our method remains consistently effective. We implement it based on BadEncoder and ResNet-18.

\paragraph{Training-free methods and inference budget.}
Supplementary Table~\ref{tab:training-free-generation} compares the performance gains brought by our fine-tuning over training-free methods. Following~\cite{ding2023clip}, we project poisoned image features into the text feature space to enable feature reconstruction without training. Fig.~\ref{fig:roc_curves_with_diff_solvers} shows the performance differences across different ODE solvers and numbers of inference steps. We find that fine-tuning yields substantial improvements, increasing the AUROC by 0.09--0.17. Moreover, 20 inference steps are already sufficient to achieve the best performance.

\subsection{Adaptive Attack}

\begin{figure*}[t]
  \begin{minipage}[t]{0.48\textwidth}
    \centering
    \captionsetup{type=figure}
    \begin{subfigure}[t]{\linewidth}
      \centering
      \includegraphics[width=\linewidth]{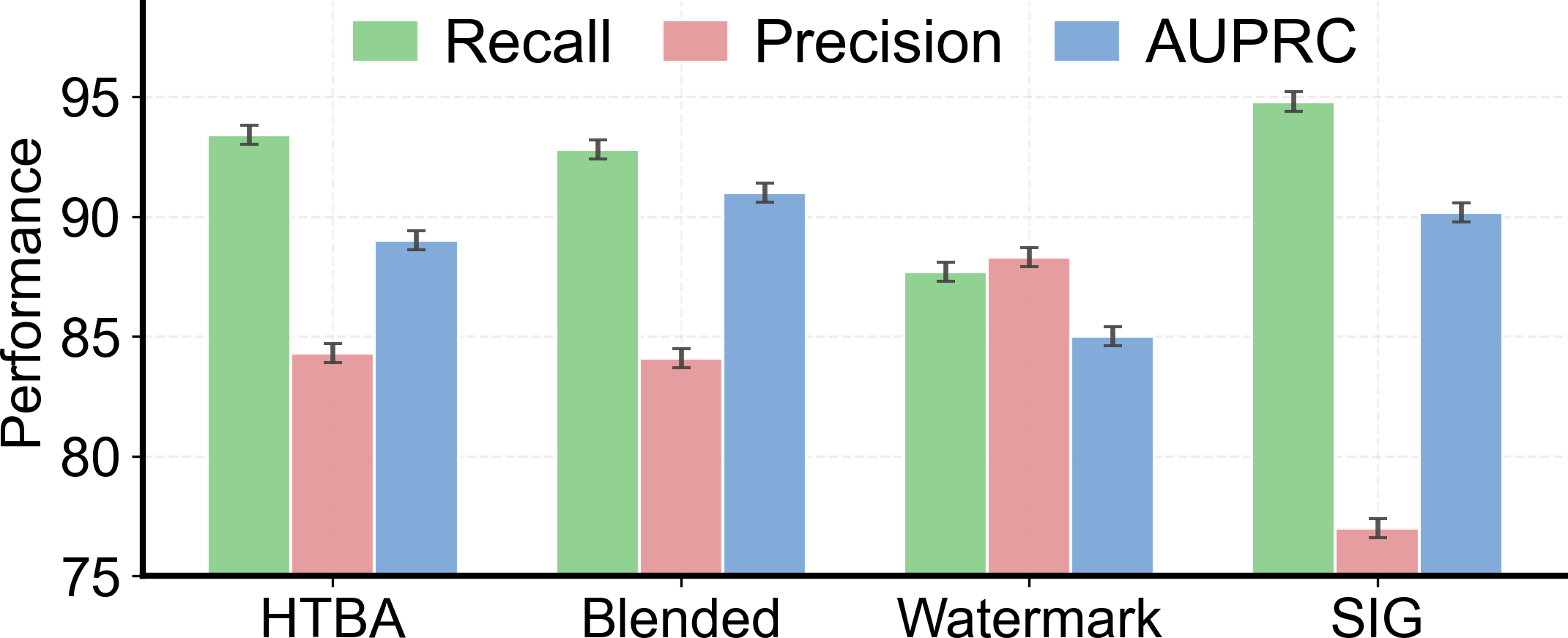}
      \caption{Performance (\%) \emph{vs.} trigger.}
      \label{fig:trigger_performance_bar}
    \end{subfigure}
    \begin{subfigure}[t]{\linewidth}
      \centering
      \includegraphics[width=\linewidth]{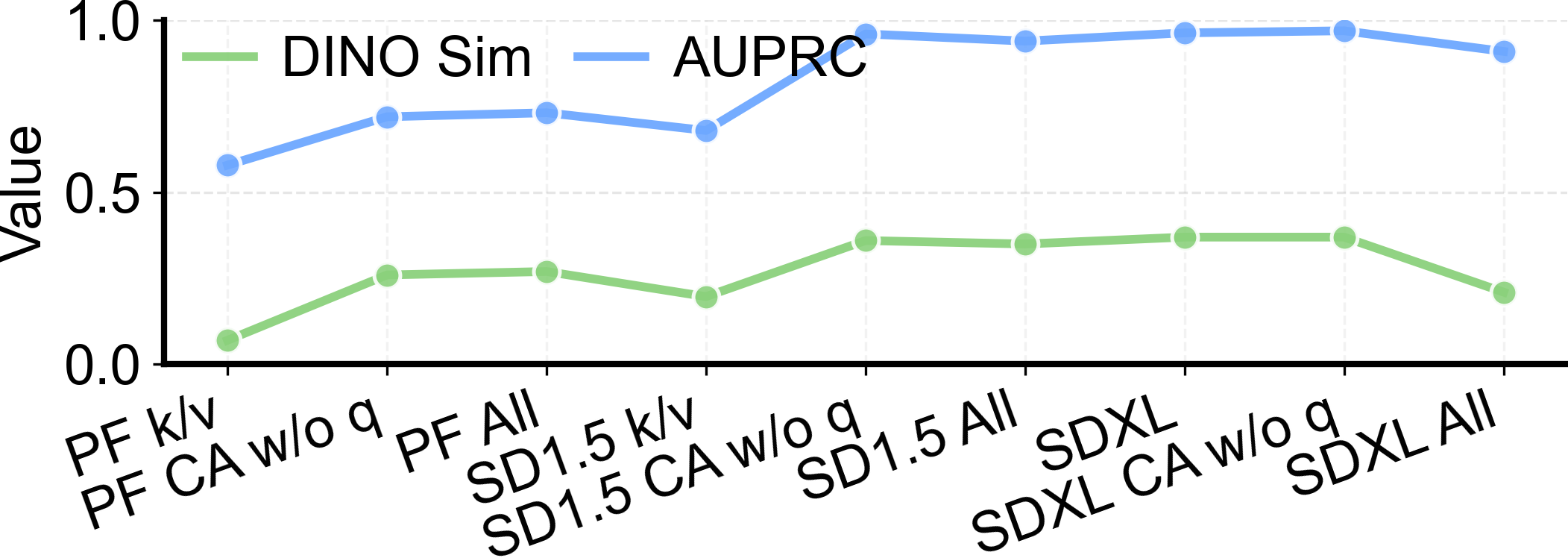}
      \caption{Trainable module ablation.}
      \label{fig:ablation_modules_lines}
    \end{subfigure}
    \caption{Performance \emph{vs.} trigger and trainable module ablation.}
    \label{fig:trigger_ablation}
  \end{minipage}\hfill
  \begin{minipage}[t]{0.48\textwidth}
    \centering
    \captionsetup{type=figure}
    \includegraphics[width=0.92\linewidth]{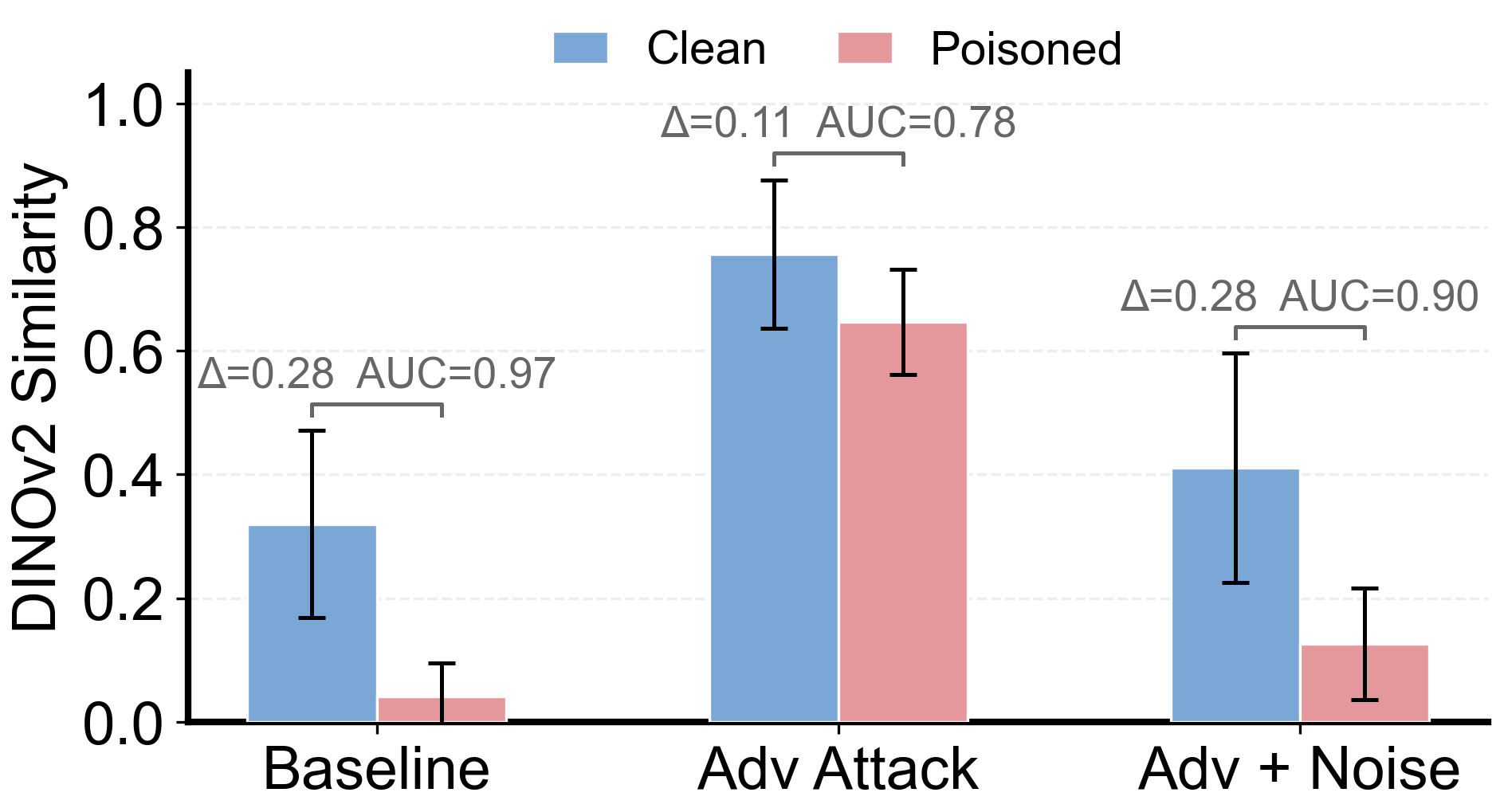}
    \caption{Semantic consistency of reconstructions under adaptive attack and defenses.}
    \label{fig:adaptive_similarity_boxplot}
  \end{minipage}
\end{figure*}

\paragraph{Adversarial attack.}
The adversary can craft adversarial perturbations to artificially increase semantic consistency. Specifically, we consider attacking Eq. \eqref{eq:dino_similarity} with PGD \cite{madry2019Deep}:
\begin{align}
    \max_{\boldsymbol{\delta}}\; \mathrm{Cosine}(\phi(\bm{x + \delta}), \phi(\epsilon_{\theta^*}(f(\bm{x+\delta}) W^*))),\,
    \text{s.t. } \|\boldsymbol{\delta}\|_{\infty}\le 16/255 ,
\end{align}
where $\delta$ denotes the adversarial perturbation. We use 20 PGD steps with a step size of $2/255$. To reduce memory consumption while preserving end-to-end gradient flow to $\bm{x}$, we employ gradient checkpointing.
Fig.~\ref{fig:adaptive_similarity_boxplot} visualizes the similarity scores of 50 randomly selected images under this adaptive attack.
Adversarial noise successfully increases reconstruction similarity and reduces the separability of backdoor images. We further consider a simple adaptive defense that adds Gaussian noise with variance $0.1$ to images carrying adversarial perturbations. This simple defense restores the AUROC from $0.78$ to $0.90$.

\paragraph{Adaptive poisoning.}
If the training data $\mathcal{D}$ still contain backdoored samples, $\mathcal{L}_{\text{z2i}}$ in Eq.~\eqref{eq:ddpm_loss} is confounded by \(P(\bm{x}\mid \mathds{1}_{s(\bm{x})}=1, \bm{z})\). We consider a setting with up to 30,000 malicious samples ($\sim$3.33\%). Fig.~\ref{fig:number_of_poisons} shows that our method is not significantly affected by adaptive poisoning. We hypothesize that this robustness arises because the pretrained generative model already imposes a strong constraint on the mapping from representations to images. As a result, even if a small number of backdoor images are mixed into the training data during fine-tuning, the model still fails to learn to reconstruct backdoored images.

%% file: sec/appe.tex
\renewcommand{\dbltopfraction}{0.95}
\renewcommand{\dblfloatpagefraction}{0.55}
\setcounter{dbltopnumber}{2}
\renewcommand{\topfraction}{0.9}
\renewcommand{\floatpagefraction}{0.55}
\setcounter{topnumber}{3}
\setcounter{totalnumber}{5}
\setlength{\dbltextfloatsep}{10pt plus 3pt minus 4pt}
\setlength{\textfloatsep}{8pt plus 2pt minus 3pt}

\section{Impact of Detection Preference}
\label{sec:analysis_of_detection_preference}

To characterize the preference of different reconstruction metrics, we analyze 150{,}000 images from CC3M and rank them by the L2 reconstruction error and the DINOv2 similarity. The qualitative examples in Figure~\ref{top_bottom_images} show that these two metrics favor markedly different image types. L2 primarily penalizes images with strong low-level complexity, whereas DINOv2 is more sensitive to whether the image contains a clear and semantically coherent foreground object.

For L2, the most difficult images typically contain large intensity contrast, strong local fluctuations, bright regions, and visually cluttered high-frequency patterns, including text, logos, icons, and fine-grained part structures. In contrast, the easiest cases are smooth scenes with large homogeneous backgrounds and weak texture variation. This trend is consistent with Table~\ref{tab:metric_explanations}: the dominant explanatory variables for L2 are the local variance standard deviation, gray-level standard deviation, bright-pixel fraction, dynamic range, and spectral entropy. In particular, L2 is strongly correlated with texture complexity, and images with heterogeneous local texture are systematically harder under this metric. The full-feature regression reaches $R^2{=}0.503$, indicating that a substantial fraction of the L2 preference can be explained by low-level image statistics.

\begin{figure}[t]
    \centering
    \includegraphics[width=0.92\linewidth]{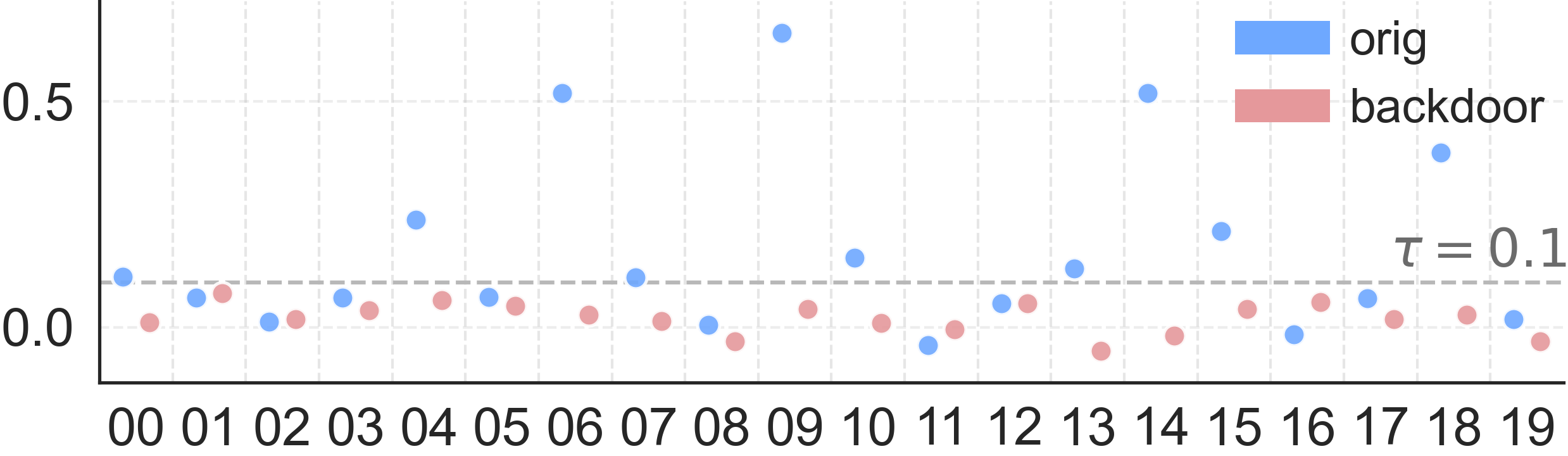}
    \caption{DINOv2 similarity of reconstructions on hard images before and after backdoor insertion.}
    \Description{Scatter plot of DINOv2 reconstruction similarity for twenty challenging images indexed from 00 to 19. Blue points show the original clean images and red points show their backdoored versions. Most clean scores lie above the threshold of 0.1, including several values above 0.5, whereas nearly all backdoored scores cluster near zero and below the threshold.}
    \label{fig:hard_images_similarity}
\end{figure}

By contrast, DINOv2 similarity tends to favor images with a single salient object, clear category semantics, limited object count, and relatively clean composition, often with the subject near the center and a sizable background region. The lowest-scoring examples are more likely to contain multiple objects, crowded scenes, ambiguous categories, fine-grained parts, or distracting textual and symbolic content. Table~\ref{tab:metric_explanations} further shows that low-level statistics explain DINOv2 similarity much less effectively, with the best full-feature model reaching only $R^2{=}0.066$. This weak explainability suggests that DINOv2 similarity is governed less by raw pixel complexity and more by semantic clarity and scene composition. Even for images selected as difficult cases according to either metric, reconstruction similarity still drops substantially after backdoor insertion, as shown in Figure~\ref{fig:hard_images_similarity}. This result indicates that the detection signal remains effective on challenging clean images.

\begin{figure*}[t]
    \centering
    \begin{subfigure}{\textwidth}
        \centering
        \includegraphics[width=\textwidth]{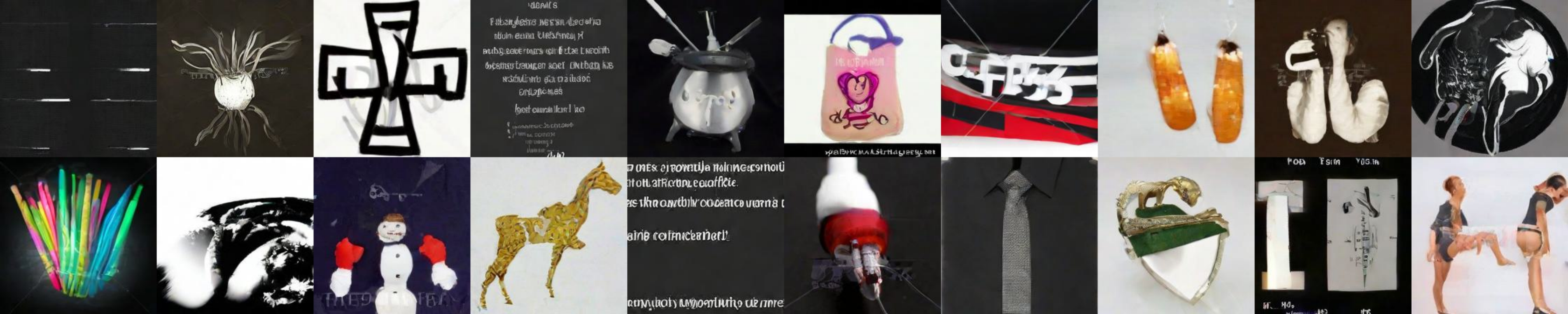}
        \caption{L2 top-20 images}
    \end{subfigure}

    \vskip 0.35em

    \begin{subfigure}{\textwidth}
        \centering
        \includegraphics[width=\textwidth]{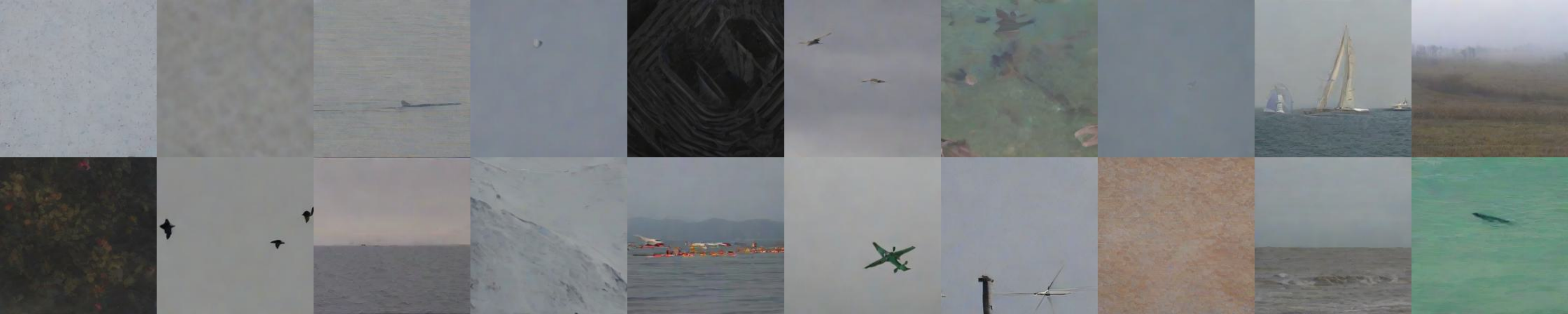}
        \caption{L2 bottom-20 images}
    \end{subfigure}

    \vskip 0.35em

    \begin{subfigure}{\textwidth}
        \centering
        \includegraphics[width=\textwidth]{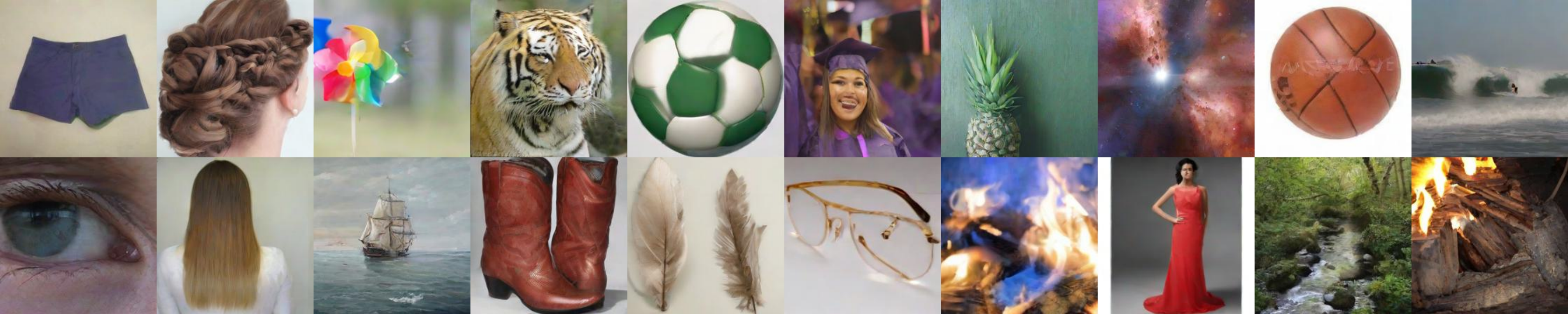}
        \caption{DINOv2 top-20 images}
    \end{subfigure}

    \vskip 0.35em

    \begin{subfigure}{\textwidth}
        \centering
        \includegraphics[width=\textwidth]{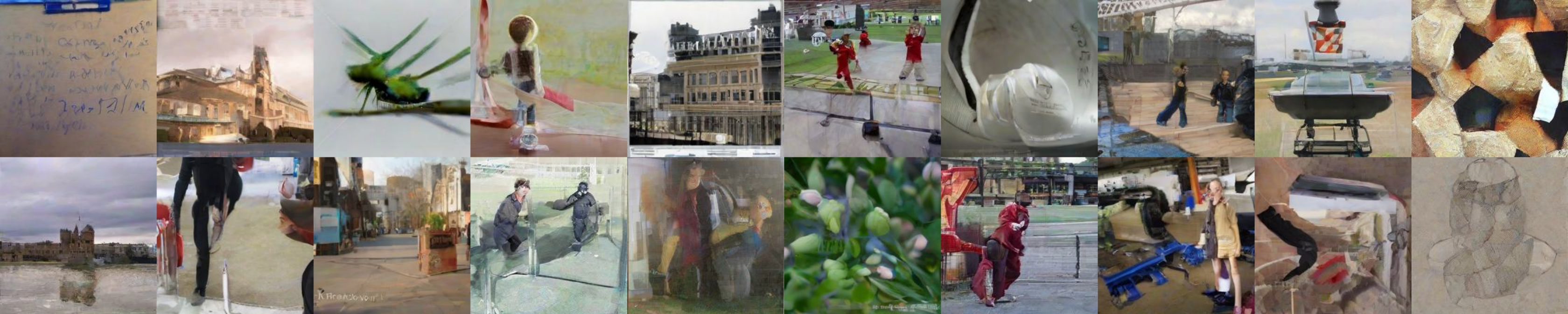}
        \caption{DINOv2 bottom-20 images}
    \end{subfigure}
    \caption{Top-20 and bottom-20 reconstructed images under the L2 and DINOv2 metrics.}
    \Description{Four grids of twenty images ranked by reconstruction metrics. Images with the largest L2 errors contain high contrast, text, icons, fine structures, or complex objects, whereas those with the smallest L2 errors are smooth, low-texture scenes such as sky, water, or fog. Images with the highest DINOv2 similarity usually contain one clear, salient object, while the lowest-ranked images contain cluttered scenes, multiple objects, ambiguous content, or text.}
    \label{top_bottom_images}
\end{figure*}

\section{Generalization of ImageNet-Trained CDDM to OOD Domains}
\label{sec:appendix-ood-generalization}

To verify that the conditional generative models learned from ImageNet remain useful beyond the natural-image domain, we evaluate the ImageNet-finetuned SDXL detector on medical and remote sensing datasets. This experiment is intended to show that the generator captures transferable semantic structure rather than overfitting to ImageNet-specific appearance statistics. As shown in Table~\ref{tab:non-natural-domains}, the detector maintains strong performance.

\section{Do Objects Matter for Caption-based Similarity?}

Vision-language models are trained on human-written descriptions of visual content and therefore naturally favor human semantics when interpreting images. In particular, their understanding is likely to be organized around objects and the relations among them, rather than only global scene context.
To verify this hypothesis, we conduct a simple intervention by modifying the Qwen3-VL-235B-A22B prompt to ``Describe the 5 main objects in the image in English'' and then recompute the Caption CLIP score. As shown in Figure~\ref{fig:objcap_distribution}, this object-focused variant improves the detection performance from a naive AUPRC of 0.904 to 0.938. This result suggests that the discriminative power of Caption-based CLIP is driven mainly by its sensitivity to object semantics, which makes backdoor samples more distinguishable from clean ones.

\section{Masked Autoencoders are not Good CDDMs}

Hou et al.~\cite{hou2025DeDe} employ masked autoencoders \cite{he2022Masked} as CDDMs. They observe that feeding only the representation is insufficient for high-fidelity reconstruction and therefore propose incorporating the extra information provided by masked autoencoders. However, masked autoencoders supply rich visual information that facilitates image reconstruction, causing the decoder to rely heavily on this information and largely ignore the information from the backdoor encoder.
We illustrate the trade-off between these two sources of encoder information through a controlled experiment on 32$\times$32-pixel GTSRB (Figure~\ref{fig:dede_trigger_size}). Different backdoor trigger sizes yield similar attack success rates, yet a larger trigger size leaves less exploitable clean information and thereby forces the decoder to rely more on the backdoor encoder's information. As the trigger size decreases, reconstruction quality improves significantly while the distinguishability of backdoor samples drops substantially. When the trigger size is 4, the reconstructed images closely resemble the originals and the detector's AUC drops to 82.4\%, whereas the attack success rate remains as high as 96\%.

\begin{figure*}[t]
\centering
\small
\setlength{\tabcolsep}{5pt}
\makeatletter\def\@captype{table}\makeatother
\caption{Explanatory variables for the two reconstruction metrics on 150,000 CC3M images.
The best full-feature model reaches $R^2{=}0.503$ for L2 but only $R^2{=}0.066$ for DINOv2 similarity.}
\label{tab:metric_explanations}
\resizebox{\linewidth}{!}{%
\begin{tabular}{l l c c l l c c}
\toprule
\multicolumn{4}{c}{L2 Error} & \multicolumn{4}{c}{DINOv2 Similarity} \\
\cmidrule(lr){1-4}\cmidrule(lr){5-8}
Feature & Category & Spearman $\rho$ & Single-feature $R^2$ & Feature & Category & Spearman $\rho$ & Single-feature $R^2$ \\
\midrule
Local variance std. & Texture & 0.549 & 0.289 & Sobel gradient std. & Edge/texture & -0.164 & 0.029 \\
Gray-level std. & Grayscale & 0.486 & 0.205 & Local variance std. & Texture & -0.149 & 0.022 \\
Bright-pixel fraction & Grayscale & 0.414 & 0.187 & Mid-frequency energy ratio & Frequency & -0.113 & 0.014 \\
Dynamic range (P95--P05) & Grayscale & 0.430 & 0.175 & Local variance mean & Texture & -0.145 & 0.022 \\
Spectral entropy & Frequency & 0.270 & 0.062 & Laplacian variance & Edge/texture & -0.117 & 0.013 \\
\bottomrule
\end{tabular}}

\vskip 0.4em

\begin{minipage}[t]{0.48\textwidth}
\vspace{0pt}
\centering
\small
\makeatletter\def\@captype{table}\makeatother
\caption{Generalization performance on medical and remote-sensing domains with ImageNet-finetuned SDXL.}
\label{tab:non-natural-domains}
\resizebox{\linewidth}{!}{%
\begin{tabular}{@{}llccc@{}}
    \toprule
    \textbf{Dataset} & \textbf{Domain} & \textbf{TPR} & \textbf{FPR} & \textbf{AUROC} \\
    \midrule
    NIH ChestX-ray14~\cite{wang2017chestx} & Med. & 0.87 & 0.17 & 0.90 \\
    Aerial Image Dataset (AID)~\cite{xia2017aid} & R.S. & 0.80 & 0.02 & 0.96 \\
    HAM10000~\cite{tschandl2018ham10000} & Med. & 0.97 & 0.08 & 0.98 \\
    NWPU-RESISC45~\cite{cheng2017remote} & R.S. & 0.89 & 0.10 & 0.95 \\
    \bottomrule
\end{tabular}}
\end{minipage}
\hfill
\begin{minipage}[t]{0.50\textwidth}
\vspace{0pt}
\centering
\makeatletter\def\@captype{figure}\makeatother
\stepcounter{figure}
\setcounter{subfigure}{0}
\begin{subfigure}{0.49\linewidth}
    \centering
    \includegraphics[width=\linewidth]{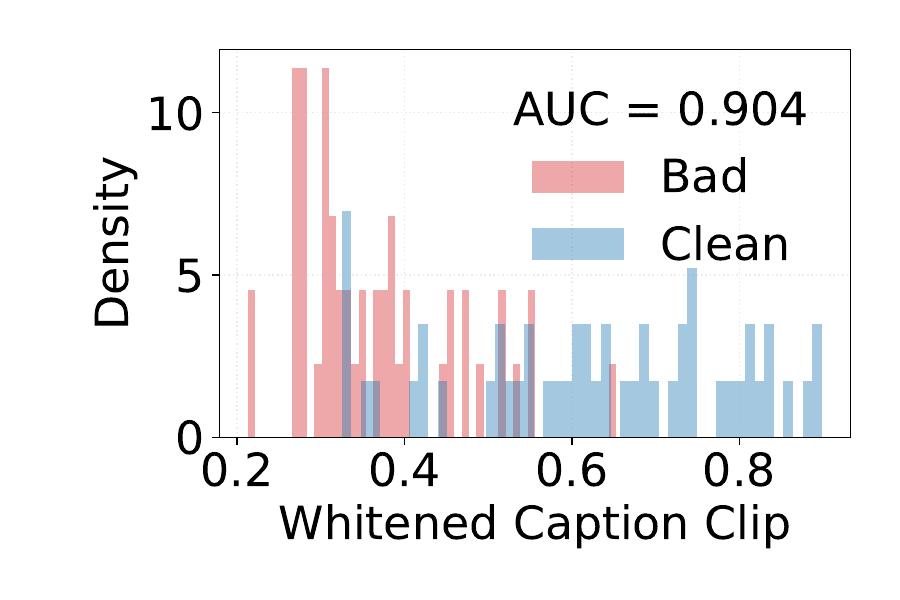}
    \caption{Overall description}
    \label{fig:overall_caption_dist}
\end{subfigure}
\hfill
\begin{subfigure}{0.49\linewidth}
    \centering
    \includegraphics[width=\linewidth]{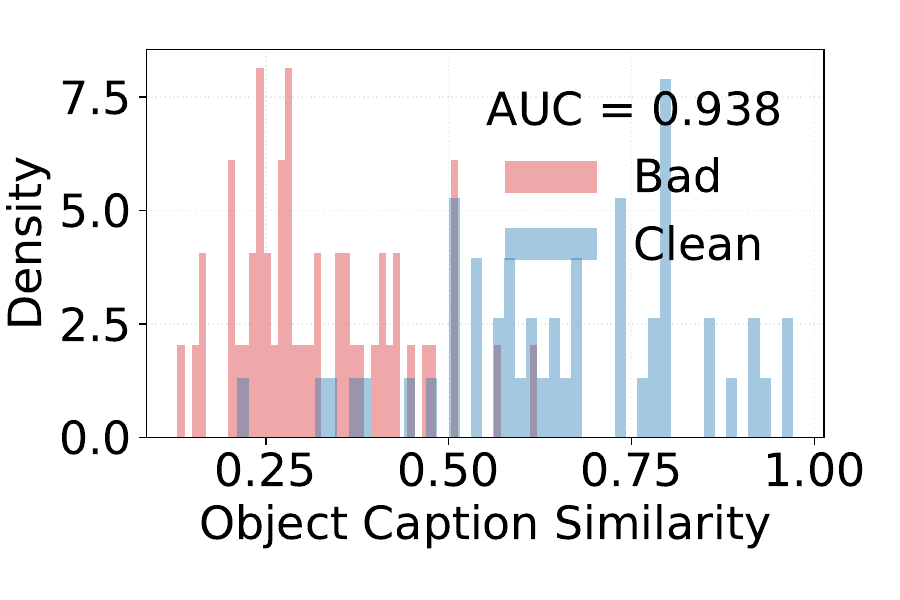}
    \caption{Object-centric description}
    \label{fig:obj_caption_dist}
\end{subfigure}
\nextfloat
\addtocounter{figure}{-1}
\caption{Distribution of Caption CLIP scores. (a) Using overall image descriptions. (b) Using object-centric descriptions.}
\Description{Two overlaid clean-versus-backdoored histograms. Using overall image descriptions, clean scores tend to be higher than backdoored scores and produce an AUC of 0.904. Using object-centric descriptions shifts the groups farther apart and increases the AUC to 0.938.}
\label{fig:objcap_distribution}
\end{minipage}

\vskip 0.35em

\includegraphics[width=0.74\textwidth]{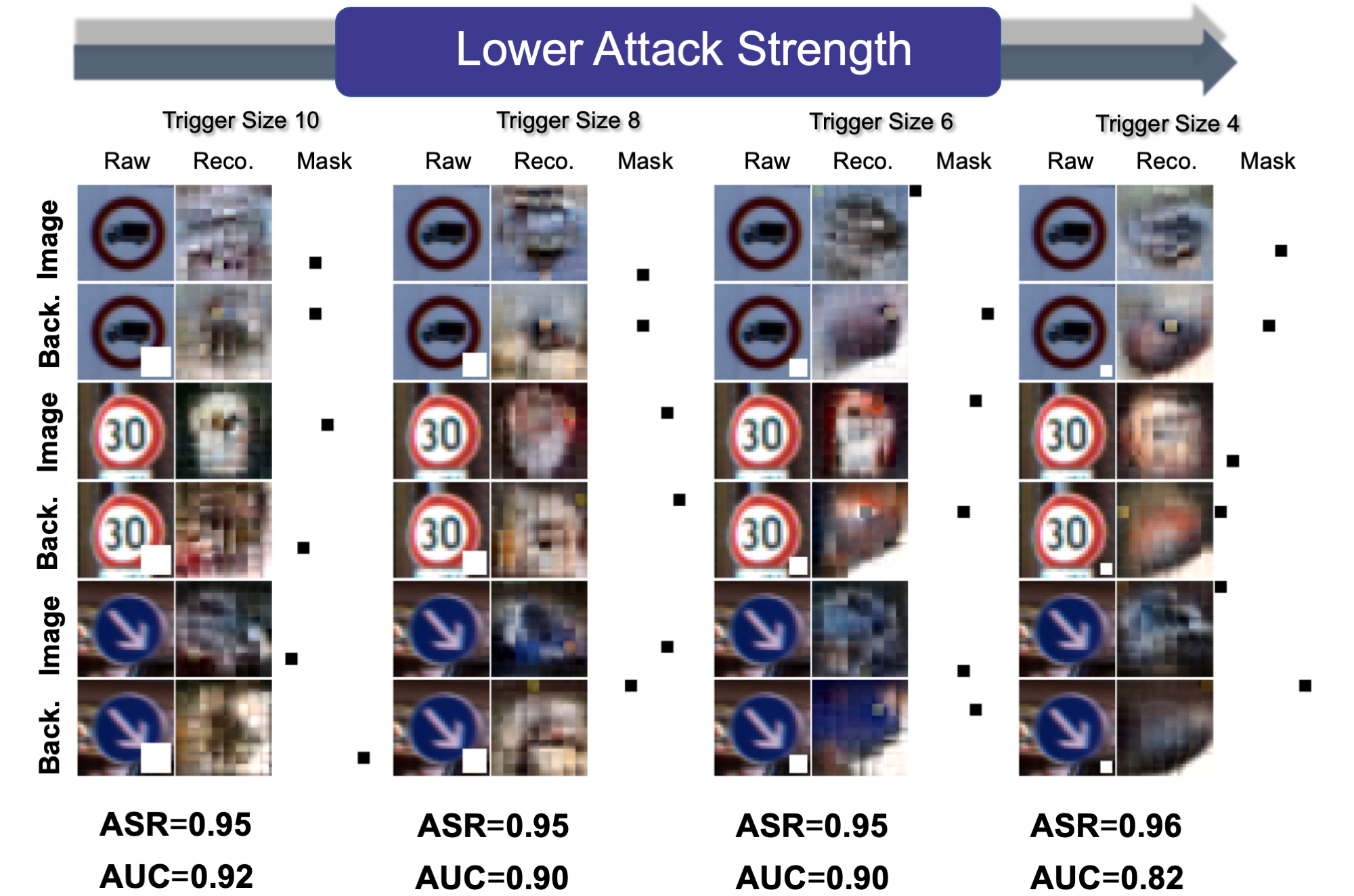}
\makeatletter\def\@captype{figure}\makeatother
\caption{Varying the trigger size of DEDE.}
\Description{Qualitative comparison for square trigger sizes 10, 8, 6, and 4 on GTSRB traffic-sign images. Each group shows clean and backdoored raw images, masked-autoencoder reconstructions, and trigger masks. As the trigger becomes smaller, attack success remains nearly constant at 0.95 to 0.96, while detection AUC decreases from 0.92 to 0.82 because the clean and backdoored reconstructions become less distinguishable.}
\label{fig:dede_trigger_size}
\end{figure*}

\section{Low-level Distances are Susceptible to the Curse of Dimensionality}

For high-dimensional images generated by a CDDM, using low-level distances (e.g., L1, L2) is inadvisable because these metrics lose sensitivity in high-dimensional spaces. Recall a classical result from high-dimensional probability~\cite{vershynin2018hdp}: for any fixed $p \ge 1$,
\begin{equation}
\frac{\|\bm{x}\|_p}{|\mathcal{X}|^{1/p}}
\;\overset{a.s.}{\longrightarrow}\;
\bigl(2^{p/2}\,\Gamma\bigl(\tfrac{p+1}{2}\bigr)/\sqrt{\pi}\bigr)^{1/p},
\end{equation}
where $\Gamma(\cdot)$ is the Gamma function and $\bm{x} \in \mathbb{R}^{|\mathcal{X}|}$ is drawn from the standard normal distribution $\mathcal{N}(\bm{0},\mathrm{I}_{|\mathcal{X}|})$. In particular,
\begin{equation}
\frac{\|\bm{x}\|_1}{\sqrt{2/\pi}\,|\mathcal{X}|}
\;\overset{a.s.}{\longrightarrow}\;1,
\qquad
\frac{\|\bm{x}\|_2}{\sqrt{|\mathcal{X}|}}
\;\overset{a.s.}{\longrightarrow}\;1.
\end{equation}
Thus, as the dimensionality increases, Euclidean distances become nearly equivalent up to constant factors.

\section{Experimental Details}

\begin{table*}[t]
\centering
\small
\begin{minipage}[t]{0.48\textwidth}
\vspace{0pt}
\centering
\footnotesize
\renewcommand{\arraystretch}{1.22}
\captionsetup{type=table,skip=4pt}
\caption{Hyper-parameters of finetuning Stable Diffusion XL.}
\label{tab:train-sdxl-hparams}
\setlength{\tabcolsep}{4pt}
\begin{tabular}{@{}l p{0.58\linewidth}@{}}
\toprule
\textbf{Category} & \textbf{Setting} \\ \midrule
\multicolumn{2}{c}{\textit{Models \& Networks}} \\ \cmidrule(lr){1-2}
Base model & Stable Diffusion XL \\
UNet & \texttt{UNet2DConditionModel} \\
Text encoder & \texttt{CLIPTextModel} \\
\multirow{2}{*}{Image encoder} & Transformer: \texttt{CLIP} \\
& CNN: \texttt{ResNet18} \\
IP-Adapter & \texttt{ImageProjModel} \\ \midrule
\multicolumn{2}{c}{\textit{Training Hyper-parameters}} \\ \cmidrule(lr){1-2}
Resolution & 512 \\
Batch size & 64 \\
Learning rate & 1e-4 \\
Optimizer & AdamW \\
Noise scheduler & DDPM \\
Mixed precision & fp16 \\
Epochs & 1 \\ \midrule
\multicolumn{2}{c}{\textit{Data \& Evaluation}} \\ \cmidrule(lr){1-2}
Training data & ImageNet-900 \\
Eval.\ data & IN-1K val.\ (image-text); IN-100 val.\ (visual SSL) \\ \bottomrule
\end{tabular}
\end{minipage}%
\hfill
\begin{minipage}[t]{0.48\textwidth}
\vspace{0pt}
\centering
\footnotesize
\renewcommand{\arraystretch}{1.22}
\captionsetup{type=table,skip=4pt}
\caption{Hyperparameters for SimSiam pretraining.}
\label{tab:visual_ssl_pretraining_config}
\setlength{\tabcolsep}{6pt}
\begin{tabular}{@{}lc@{}}
  \toprule
  Methods & SimSiam \\
  \midrule
  Training Epochs & 300 \\
  Batch Size & 512 \\
  Optimizer & SGD \\
  Learning Rate Schedule & Cosine \\
  Learning Rate & 0.06 \\
  Weight Decay & \(1 \times 10^{-4}\) \\
  Moving Average & 0.999 \\
  \midrule
  Resize \& Crop & RandomResizeAndCrop \\
  Color Jitter & 0.4 \\
  RandomHorizontalFlip & 0.5 \\
  Min Crop Scale & 0.2 \\
  RandomGrayscale & 0.2 \\
  GaussianBlur(p=0.5) & [.1, 2.] \\
  \bottomrule
\end{tabular}

\vskip 0.7em

\small
\renewcommand{\arraystretch}{1.05}
\captionsetup{type=table,skip=4pt}
\caption{Training-free \cite{ding2023clip} vs.\ finetuned.}
\label{tab:training-free-generation}
\setlength{\tabcolsep}{4pt}
\begin{tabular}{@{}lcccc@{}}
    \toprule
    \textbf{Metric} & \textbf{TPR} & \textbf{FPR} & \textbf{AUROC} & \textbf{FT.\ AUROC} \\
    \midrule
    DINOv2 & 83.0\% & 19.0\% & 0.86 & 0.95 \\
    ViT & 76.0\% & 29.0\% & 0.76 & 0.93 \\
    \bottomrule
\end{tabular}
\end{minipage}

\vskip 0.7em

\captionsetup{type=table,skip=4pt}
\caption{Downstream Performance.}
\label{tab:downstream}
\renewcommand{\arraystretch}{1.08}
\setlength{\tabcolsep}{3pt}
\resizebox{\linewidth}{!}{%
\begin{tabular}{l ccc ccc ccc ccc}
    \toprule
    & \multicolumn{3}{c}{\textbf{SSLBKD}}
    & \multicolumn{3}{c}{\textbf{BadEncoder}}
    & \multicolumn{3}{c}{\textbf{CLIP Backdoor}}
    & \multicolumn{3}{c}{\textbf{BadCLIP}} \\
    \cmidrule(lr){2-4}\cmidrule(lr){5-7}\cmidrule(lr){8-10}\cmidrule(lr){11-13}
    & R. (\%) & CA (\%) & ASR (\%)
    & R. (\%) & CA (\%) & ASR (\%)
    & R. (\%) & CA (\%) & ASR (\%)
    & R. (\%) & CA (\%) & ASR (\%) \\
    \midrule
    No Defense & - & 65.1 & 51.2 & - & 62.0 & 84.1 & - & 62.9 & 95.2 & - & 60.1 & 88.9 \\
    PatchSearch & 23.1 & 65.2 & 58.1 & 27.9 & 61.8 & 91.0 & 11.5 & 63.3 & 87.0 & 8.7 & 63.1 & 85.2 \\
    \hspace{0.5em}{\textbf{\texttt{+}}} Classifier & 98.3 & 66.5 & 0.9 & 99.9 & 61.2 & 1.2 & 26.9 & 60.6 & 93.0 & 17.8 & 63.3 & 76.2 \\
    DBCL & 8.5 & 66.3 & 70.9 & 13.6 & 59.6 & 93.4 & 93.8 & 60.9 & 2.1 & 19.5 & 63.6 & 84.4 \\
    DEDE & 12.2 & 66.1 & 68.7 & 15.9 & 63.2 & 93.7 & 16.0 & 62.4 & 88.3 & 9.3 & 64.0 & 87.3 \\
    \cellcolor{lightgray}DEFUSE & \cellcolor{lightgray}43.8 & \cellcolor{lightgray}65.9 & \cellcolor{lightgray}49.1 & \cellcolor{lightgray}39.4 & \cellcolor{lightgray}61.1 & \cellcolor{lightgray}87.9 & \cellcolor{lightgray}92.6 & \cellcolor{lightgray}63.9 & \cellcolor{lightgray}0.0 & \cellcolor{lightgray}73.4 & \cellcolor{lightgray}61.8 & \cellcolor{lightgray}73.1 \\
    \cellcolor{lightgray}\hspace{0.5em}{\textbf{\texttt{+}}} Classifier & \cellcolor{lightgray}99.8 & \cellcolor{lightgray}65.8 & \cellcolor{lightgray}0.1 & \cellcolor{lightgray}99.9 & \cellcolor{lightgray}61.7 & \cellcolor{lightgray}0.1 & \cellcolor{lightgray}100.0 & \cellcolor{lightgray}63.9 & \cellcolor{lightgray}0.1 & \cellcolor{lightgray}99.9 & \cellcolor{lightgray}62.2 & \cellcolor{lightgray}0.0 \\
    \bottomrule
\end{tabular}}
\end{table*}

Table~\ref{tab:train-sdxl-hparams} summarizes the training hyper-parameters and experimental settings for Stable Diffusion XL.
We implement the code following IP-Adapter \cite{ye2023IPAdapter} and the Diffusers library from Hugging Face. Unless otherwise specified, for CLIP experiments we use the HTBA patch trigger as the default backdoor pattern, with a trigger size $50\times50$. We fine-tune the poisoned CLIP model for 10 epochs with a learning rate $1\times10^{-4}$ and a batch size 256. For blended-trigger experiments, we set the transparency coefficient $\alpha$ to 0.2. We evaluate the ASR of CLIP under the zero-shot protocol~\cite{radford2021Learning}.
For ResNet18, we first align the feature dimension to match the pretrained CDDM. In the visual SSL setting, we pretrain SimSiam-ResNet18 for 300 epochs and otherwise follow the official SimSiam optimization setup~\cite{chen2021Exploring}. We also use an HTBA trigger of size $50\times50$ and evaluate ASR using a linear-probe protocol. We reuse the same ImageNet-100 class split as in~\cite{saha2022Backdoor} for the attack and evaluation pipeline, while using the remaining 900 ImageNet-1K classes for CDDM fine-tuning, thereby preventing the attack data from being inadvertently exploited during generative model adaptation.
Table~\ref{tab:visual_ssl_pretraining_config} summarizes the experimental setup for visual self-supervised pretraining.